\documentclass[lettersize,journal]{IEEEtran}
\usepackage{amsmath,amsfonts}
\usepackage{algorithmic}
\usepackage{array}
\usepackage[caption=false,font=normalsize,labelfont=sf,textfont=sf]{subfig}
\usepackage{textcomp}
\usepackage{stfloats}
\usepackage{url}
\usepackage{verbatim}
\usepackage{graphicx}
\def\BibTeX{{\rm B\kern-.05em{\sc i\kern-.025em b}\kern-.08em
    T\kern-.1667em\lower.7ex\hbox{E}\kern-.125emX}}
\usepackage{balance}

\usepackage{soul}
\usepackage{xcolor}
\usepackage{subcaption}
\usepackage{multirow}
\usepackage{booktabs}
\usepackage{tabularx}
\usepackage{caption}
\usepackage[most]{tcolorbox}

\newtcbox{\roundhl}[1][yellow]{%
    on line, arc=2mm, colback=#1!30!white, colframe=#1!80!black,
    before upper=\strut, boxsep=0pt, left=2pt, right=2pt, top=1pt, bottom=1pt,
    boxrule=0.2pt
}
\makeatletter
\newcommand\footnoteref[1]{\protected@xdef\@thefnmark{\ref{#1}}\@footnotemark}
\makeatother

\usepackage[hidelinks]{hyperref}

\begin{document}
\title{SocialMirror: Body Behavior Recognition in Adults with Intellectual and Developmental Disabilities}

\author{Simona Mastroberardino$^{1}$,~\IEEEmembership{}
        Qi Ai$^{1}$,~\IEEEmembership{}
        Maristella Matera$^{1}$,~\IEEEmembership{}
        Micol Spitale$^{1}$~\IEEEmembership{}
\thanks{$^1$ Politecnico di Milano, Italy; contact author email: micol.spitale@polimi.it}
}

\markboth{Journal of \LaTeX\ Class Files,~Vol.~18, No.~9, September~2020}%
{How to Use the IEEEtran \LaTeX \ Templates}

\maketitle

\begin{abstract}
AI systems for human behavior recognition are typically trained on datasets representing predominantly neurotypical populations, limiting their inclusivity and applicability in care settings. At the same time, in daycare centers, educators for individuals with intellectual and developmental disabilities (IDD) face continuous monitoring demands, making it difficult to detect subtle non-verbal cues of distress or potential harm. 
To address this gap, we present SocialMirror, an empirical study and multi-view body-behavior dataset collected in a naturalistic daycare setting involving individuals with intellectual and developmental disabilities (IDD) over 4 days, capturing activities of 13 participants with IDD using two camera views. Through a co-design process with educators, we developed an annotation codebook capturing both individual behaviors (e.g., closed posture, repetitive actions) and relational cues (e.g., affective interactions). The dataset consists of skeleton-based representations segmented into 2-second windows, from the two camera views and 3D reconstruction, yielding approximately 232 minutes of aggregated annotated behaviour duration.
We evaluate skeleton-based baselines on a four-class individual-level recognition task and a binary pair-level relational recognition task.
Our findings highlight four challenges and we draw on these findings to formulate methodological lessons for \textit{inclusive} affective computing.
\end{abstract}


\section{Introduction}
Recent advances in artificial intelligence for human behavior understanding have shown promising results in domains such as action recognition \cite{gupta2022human}, activity monitoring \cite{hussain2024ai}, and assistive technologies \cite{zdravkova2022potential}. However, these systems are typically developed and evaluated on datasets that predominantly represent neurotypical populations, limiting their inclusivity and applicability in real-world care contexts \cite{huang2026inclusive, almufareh2024conceptual}. As a result, current AI models often fail to capture the nuanced and diverse non-verbal behaviors exhibited by individuals with intellectual and developmental disabilities (IDD), raising concerns about fairness, generalization, and practical utility in inclusive settings, as demonstrated in \cite{huang2026inclusive}.
These limitations highlight the growing need for more inclusive and human-centered AI approaches \cite{barbareschi2026inclusive, shneiderman2022human}. Human-centered AI emphasizes the design of systems that are grounded in the needs, values, and contexts of end users, particularly in sensitive domains such as healthcare and assistive technologies \cite{shneiderman2022human}. Recent work has advocated for participatory and co-design methodologies to ensure that AI systems are not only technically effective but also socially meaningful and ethically aligned \cite{almufareh2024conceptual, panigutti2023co}. In the context of behavior understanding, this entails capturing socially relevant signals, incorporating domain expertise, and accounting for variability in human behavior that extends beyond standard benchmarks. 
Despite these efforts, there remains a lack of works that reflect real-world care environments and the lived experiences of individuals with IDD, limiting progress toward truly inclusive behavior recognition systems.

At the same time, care environments such as daycare centers for individuals with IDD face significant operational challenges \cite{DBLP:journals/mta/GelsominiSG21, DBLP:journals/ijmms/GennariMMMMR24}. 
Educators are required to continuously monitor multiple attendees, often without the possibility of taking breaks, due to concerns about safety and well-being \cite{DBLP:journals/mta/GelsominiSG21}. 
Discussions with educators, conducted as part of this study, highlighted that they also need to identify moments of distress, discomfort, or potentially harmful situations, which may be conveyed through subtle non-verbal cues and social interactions that are difficult to track at scale.
Increasing representation in training data addresses only one aspect of inclusion. Affective computing also depends on decisions about which behaviors are meaningful, how observations are translated into labels, which bodies and interactions can be reliably sensed, and what constitutes successful generalization. In disability contexts, these decisions require particular attention to individual communication practices and the circumstances in which behavior occurs. 

Motivated by these challenges, we take a first step toward inclusive and socially-aware AI systems by investigating body-behavior recognition as a situated methodological problem, examining the adjustments suggested by its application in a naturalistic setting involving adults with IDD.
We introduce \textit{SocialMirror}, a \textbf{dataset} for body behavior recognition in daycare activities involving individuals with IDD. The dataset was collected in a real-world daycare center over four days, capturing daily activities between 09:00 and 14:00 using two camera viewpoints, and includes 13 participants. Our goal was to capture meaningful body behaviors at both the individual and relational levels.
A key aspect of this work is the development of an \textbf{annotation framework} through a co-design process with educators from the center. This resulted in an extensive behavior codebook that reflects domain expertise and practical relevance, covering both individual behaviors (e.g., closed posture, repetitive actions) and relational cues (e.g., expressions of affection among individuals). 
We curate the dataset as skeleton-based representations and segment it into temporal samples using a 2-second windowing scheme from two camera views, yielding a total of 17,372 (camera 1), 11,580 (camera 2), and 607 (3D) annotated samples across four target classes.
We also \textbf{benchmarked} SocialMirror using complementary baseline approaches, namely a Support Vector Machine on interpretable pose-derived features and exploratory PoseC3D experiments on skeleton-based representations, formulating the \textit{individual} recognition task as a 4-class classification problem (classes: Stand, Sit, ClosedPosture, DisengagedRepetitive) across multiple camera-view and 3D-derived settings. 
In parallel, we formulate \textit{relational} behavior recognition as a pair-level binary classification task over two aggregated categories: SocialOrientation and AffiliativeContact. A total of 5,709 usable pair-level windows were available across the two camera-view datasets; to mitigate the strong class imbalance, the relational benchmark was balanced to 2,098 windows, with 1,049 instances per class, and evaluated using video-level grouped splits.
This work makes three contributions: 
\begin{itemize}
    \item we provide a situated \textbf{empirical investigation of body-behavior recognition in a naturalistic daycare setting involving adults with IDD}, combining educator-informed annotation with individual- and relational-level recognition benchmarks;
    \item we derive \textbf{methodological lessons for inclusive affective computing} from educator-informed label construction, participant--behavior distributions, pair-level observability, and diagnostic benchmark results;
    \item we \textbf{release} \textbf{SocialMirror}, comprising an annotation codebook, derived-feature datasets, and baseline evaluations to support reproducible investigation of these challenges.
\end{itemize}

\section{Related Work}

\subsection{Human-Centered AI in Inclusive Settings}
Human-Centered AI (HCAI) promotes the design of intelligent systems that are aligned with human needs, values, and real-world contexts, with the goal of enhancing human well-being, autonomy, and fairness \cite{shneiderman2022human}. This paradigm has become particularly important in domains such as healthcare and assistive technologies, where system decisions directly impact vulnerable users. To achieve these goals, HCAI emphasizes participatory and co-design approaches, actively involving end users and domain experts in the development process \cite{panigutti2023co, almufareh2024conceptual}. Recent frameworks for AI-enabled assistive technologies further highlight the importance of integrating usability, human rights, and user experience into system evaluation \cite{silvera2025ai}. In this context, approaches such as co-design in the middle have been proposed to better account for diverse cognitive and social needs, particularly by incorporating the perspectives of practitioners and marginalized communities \cite{bircanin2025co}.
However, a significant gap remains in the representation of people with disabilities within HCAI research and practice. Current AI systems are still largely trained on data from neurotypical populations, limiting their ability to generalize to neurodivergent groups such as individuals with IDD \cite{shams2025ai, xiao2024systematic}. This lack of diversity in training data creates systematic blind spots, leading to reduced performance in tasks such as speech and behavior recognition when applied to individuals with nonstandard communication patterns. As a result, these populations are often excluded from fully benefiting from digital technologies \cite{cibrian2025limitations}. Addressing this representation gap is therefore essential to realizing the promise of truly inclusive human-centered AI systems.

\subsection{Body Behaviour Modeling and Datasets}
Human behavior recognition has increasingly shifted from RGB-based representations to skeleton-based modeling, which offers advantages in robustness to background noise and improved privacy preservation \cite{liu2025systematic}. This transition is particularly relevant in sensitive contexts such as care environments, where visual appearance data may raise ethical concerns. 
Recent advances in skeleton-based action recognition have focused on improving the modeling of spatial-temporal dynamics. Graph Convolutional Networks (GCNs), such as ST-GCN \cite{yan2018spatial} and its variants \cite{lasri2025dpca}, explicitly capture the topology of human joints, while alternative approaches like PoseConv3D \cite{duan2022revisiting} leverage volumetric representations to enhance robustness. More recently, Transformer-based architectures (e.g., SkateFormer \cite{do2024skateformer}, SkeleTR \cite{duan2023skeletr}) have demonstrated strong performance in modeling long-range dependencies. Additional work has addressed real-world challenges such as occlusions, incomplete observations, and generalization to unseen behaviors \cite{shao2026finetec, wu2025frequency}. Importantly, emerging approaches such as AXON \cite{shah2026axon} show that skeleton-based representations can effectively capture atypical behavioral patterns in neurodivergent populations.
Despite these modeling advances, progress is fundamentally constrained by the availability of suitable datasets. While large-scale benchmarks and recent multi-person interaction datasets (e.g., DUET \cite{lin2025actions}, Inter-X \cite{xu2024inter}) enable the study of social behaviors, they are primarily collected in controlled or generic settings and do not reflect the complexity of care environments. In parallel, a growing body of work has started to explore datasets involving people with disabilities, such as MMASD \cite{li2023mmasd} and recent skeleton datasets for autism-related behaviors \cite{shah2026axon}, as well as benchmarks targeting safety-critical actions \cite{garcia2026summary}. 
However, these efforts remain limited in scope. Existing datasets are often restricted to laboratory conditions, short-term recordings, or highly specific interaction scenarios, and they rarely capture continuous, naturalistic behavior in real-world care settings. Moreover, the collection of such data is challenged by strict privacy constraints and the heterogeneity of behavioral patterns in individuals with IDD \cite{li2023mmasd, shah2026axon}.

These limitations highlight the need for ecologically valid, privacy-preserving datasets and benchmarks that capture both individual and social behaviors in real care environments. To address this gap, we introduce the \textit{SocialMirror}, which provides multi-view skeleton-based recordings of activities in a daycare center for individuals with IDD, annotated through a co-design process with domain experts and benchmark with individual and relational recognition tasks.

\vspace{-0.2cm}
\section{The \textit{SocialMirror} Dataset}
Prior work has highlighted educators’ need for support in monitoring social and behavioral dynamics in these settings \cite{DBLP:journals/ijmms/GennariMMMMR24, DBLP:journals/mta/GelsominiSG21}. 
Existing datasets and models were found to be inadequate to capture the variability and ecological complexity of real-world care settings involving neurodivergent populations \cite{huang2026inclusive, barbareschi2026inclusive}.
To address this gap, we collected a dataset capturing naturalistic interactions and behaviors in a real daycare center attended
by adults with IDD, named \textit{SocialMirror}. 


\subsection{Dataset Collection}
The dataset was collected through recordings capturing participants’ daily activities in a common indoor space using a multi-view camera setup, designed to unobtrusively observe natural behavior. 

\subsubsection{Participants}
We involved all attendees of the daycare center in the study. Participants were introduced to the research project by the center’s educators and provided verbal assent, while informed consent was obtained from their legal guardians. Participants were informed of their right to withdraw from the study at any time during data collection. The study was approved by the Ethical Committee of the University Politecnico di Milano.
The dataset includes 13 participants with IDD. Based on visual observations of the recordings, two independent researchers classified 8 participants as female-presenting and 5 as male-presenting, with full agreement. These classifications reflect external perceptions of gender presentation and should not be interpreted as participants' self-identified gender.
5 educators appeared in the recordings while working, but their data were discarded and not included in the dataset. 

\subsubsection{Setup}
\begin{figure}[t]
    \centering
    \subfloat[View from Camera 1 (Camera 2 visible in the scene).%
    \label{fig:cam1}]{%
        \includegraphics[width=0.9\columnwidth]{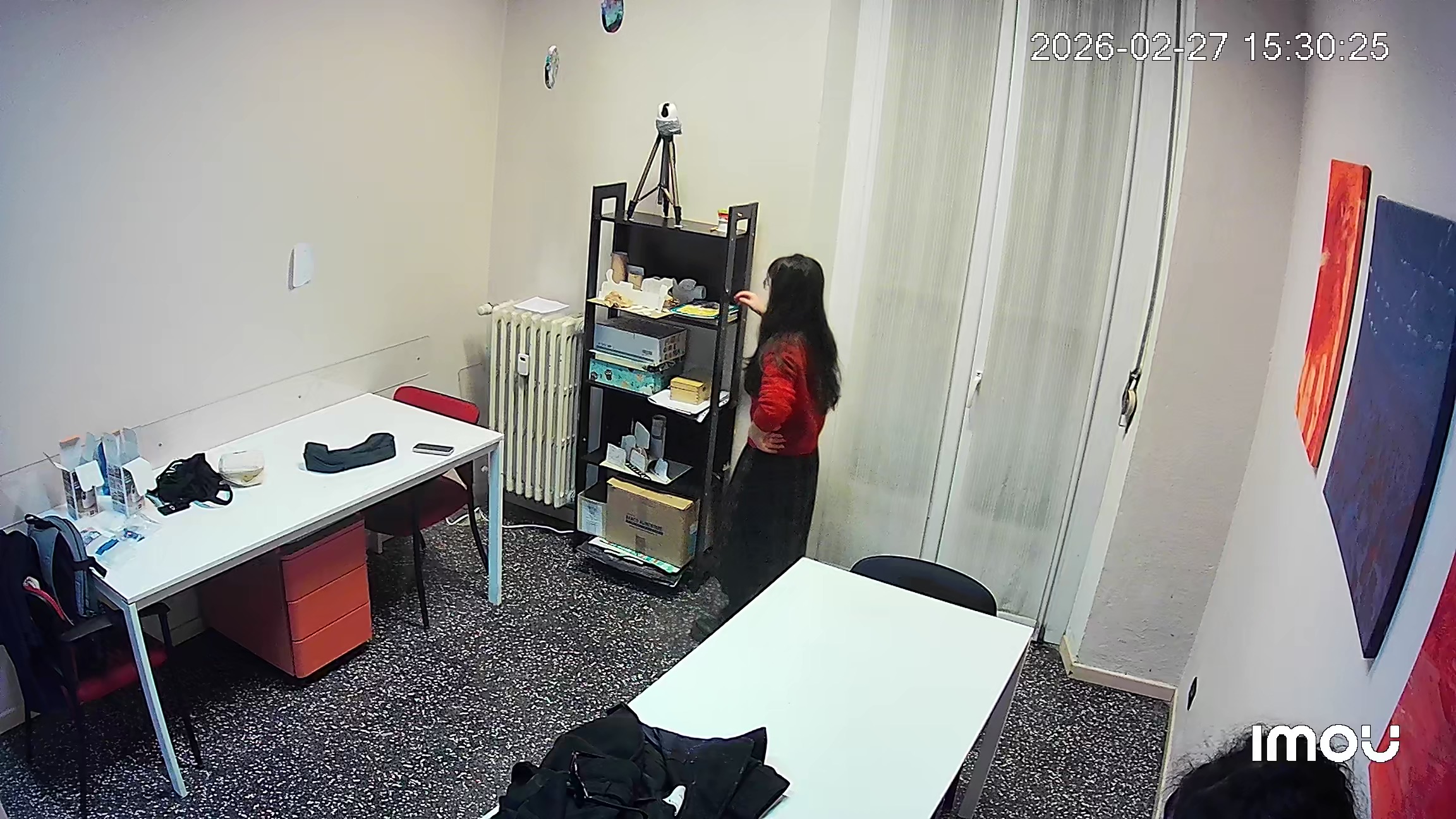}%
    }
    \vspace{0.15cm}
    \subfloat[View from Camera 2 (Camera 1 visible in the scene).%
    \label{fig:cam2}]{%
        \includegraphics[width=0.9\columnwidth]{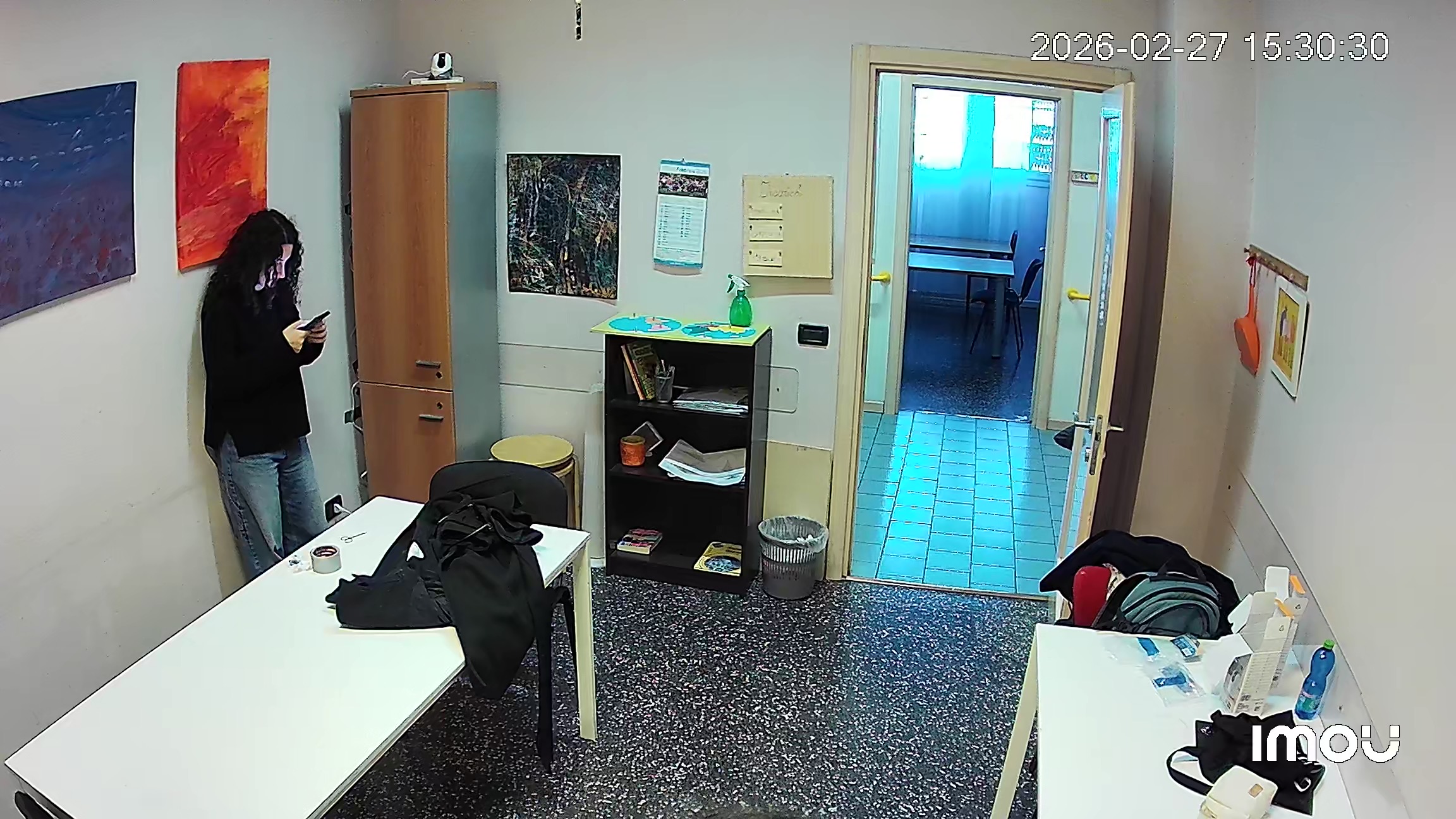}%
    }
    \vspace{-0.2cm}
    \caption{Setup of data recordings and camera positioning.}
    \label{fig:setup}
\end{figure}

We installed two Imou Ranger cameras\footnote{\url{https://www.imou.com}}
 (2C 2K, 1296p / 25fps, H.265, recordings stored on a local microSD) in opposite corners of the room to maximize spatial coverage and capture the full activity area from complementary viewpoints. These cameras were preferred over more complex recording sensors because they offered an adequate trade-off between video quality, ease of deployment, limited intrusiveness, and ecological feasibility in a real-world care setting. This configuration also ensured overlapping fields of view, enabling multi-view analysis and providing common reference points for potential triangulation.

Both cameras were positioned at elevated locations to reduce occlusions and improve visibility of participants preserving their privacy and avoiding capturing their faces directly, see Figures \ref{fig:cam1} and \ref{fig:cam2}. Camera 1 (\textit{cam1}) was placed on top of a cabinet near the entrance, while Camera 2 (\textit{cam2}) was mounted on a tripod positioned on a shelf, allowing for a slightly higher viewpoint and a broader perspective of the room. As shown in Figure~\ref{fig:setup}, each camera captures not only the activity space but also the position of the other camera, facilitating spatial alignment.
The recording environment consisted of a typical activity room, including tables, chairs, and shelves, where participants engaged in their daily routines. 

\subsubsection{Camera Calibration}
Prior to data collection, we performed camera calibration using a standard checkerboard pattern\footnote{\label{checkerboard}\url{https://docs.opencv.org/4.x/dc/dbb/tutorial_py_calibration.html}}. This procedure allowed us to estimate the intrinsic parameters of each camera (e.g., focal length, principal point, and lens distortion coefficients), which were subsequently used to correct image distortion during preprocessing.
In addition, the checkerboard was used as a shared reference object visible from both cameras to facilitate spatial alignment across views as done previously in \cite{liao2024mvgformer,chen20243dsa}. This enabled the establishment of a common coordinate frame, supporting multi-view consistency and potential triangulation of observations.
To ensure stability of the calibration, we fixed the camera positions and orientations throughout the entire recording period. Visual markers were placed in the environment to indicate the exact placement of each camera, preventing accidental displacement and preserving the calibration setup across all recording sessions.

\subsubsection{Protocol}
Data collection was designed to be as unobtrusive as possible in order to capture naturalistic behaviors during participants’ daily routines. After installing and calibrating the cameras, recordings were conducted continuously during regular daycare activities without introducing any scripted tasks or interventions.
Each recording session took place between 09:00 and 14:00, from Monday to Thursday (March 2026), covering a variety of everyday activities such as free time, group interactions, and individual tasks. Educators and participants were instructed to behave as usual, and no specific behaviors were elicited. This allowed us to capture spontaneous non-verbal behaviors in a realistic setting.

We adopted a privacy-preserving data processing pipeline \cite{moon2023anonymization, gao2024privacy}: no face recognition, biometric identification, or appearance-based features were used at any stage of the pipeline. Instead, all downstream processing relies exclusively on geometric and temporal signals (e.g., body pose and motion). In multi-view settings, cross-camera association is performed solely using these signals, thereby avoiding the use of identity-related visual cues.
Participants and educators were informed about the recording schedule, and consent procedures were completed prior to data collection. Cameras remained fixed throughout the study, and no adjustments were made during recording sessions.
At the end of each day, the recorded data were securely stored and checked for quality. Segments affected by severe occlusions, technical issues, or participant absence were flagged for exclusion during the curation process.

\subsection{Feature Extraction}

\subsubsection{Video Pre-processing}
The recorded videos were initially converted from the native \texttt{.dav} format to \texttt{.mp4} to ensure compatibility with standard processing tools. Subsequently, we applied lens distortion correction to all recordings. 
To this end, we estimated the intrinsic camera parameters and distortion coefficients using a standard camera calibration procedure based on a checkerboard pattern\footnoteref{checkerboard}. Multiple images of the checkerboard were captured from different viewpoints, allowing the extraction of correspondences between known 3D points on the calibration pattern and their 2D projections in the image plane. Using these correspondences, we computed the camera matrix (including focal lengths and principal point) and radial and tangential distortion coefficients following the OpenCV calibration pipeline.
The estimated parameters were then used to undistort all video frames, effectively correcting lens-induced geometric distortions, particularly near image boundaries. This step ensures geometrically consistent observations across views and provides rectified inputs for subsequent multi-view processing and feature extraction as done in \cite{liao2024mvgformer,chen20243dsa}.

\subsubsection{Feature Extraction}
We extract human pose representations using the MediaPipe Python library\footnote{\url{https://ai.google.dev/edge/mediapipe/solutions/guide}}, which provides real-time multi-person pose estimation based on 2D body landmarks. Each detected individual is represented by 33 keypoints covering head, upper limbs, trunk, and lower limbs. In our pipeline, tracking and filtering rely on a reduced set: a 5-point core subset (nose, left/right shoulders, left/right elbows) and a 9-point upper-body subset (nose, left/right ears, shoulders, elbows, and wrists).
Due to the unconstrained environment, initial extraction produced fragmented and unstable tracks, caused by occlusions, seated positions, and close interactions. This resulted in identity inconsistencies and short, noisy trajectories, hindering alignment with manual annotations.

To address this, we refined the pipeline in three ways. First, we improved temporal tracking to favor longer, consistent trajectories, filtering out short-lived tracks. Second, we applied quality-based filtering, removing poses with low confidence (upper-body completeness score $< 0.60$, computed with landmark visibility threshold $0.50$) or insufficient visibility (fewer than 2 visible core landmarks). Third, we adapted the pipeline to retain informative partial skeletons, given that participants were often visible only in the upper body.
The resulting output consists of temporally consistent skeletal representations capturing posture and motion, which are used for annotation alignment and downstream modeling. 
Concretely, the resulting 2D branch supported the construction of the final person-level datasets used for baseline modeling. After multi-video merging, the \textit{cam1} dataset contained 42,042 samples in the all\_windows version and 17,372 samples in the 2s\_only version, while the \textit{cam2} dataset contained 33,622 and 11,580 samples, respectively. Since the 2s\_only configuration proved more stable in the baseline experiments, this subset became the main reference for subsequent modeling.

\begin{figure}[t]
    \centering

    \includegraphics[height=3.8cm]{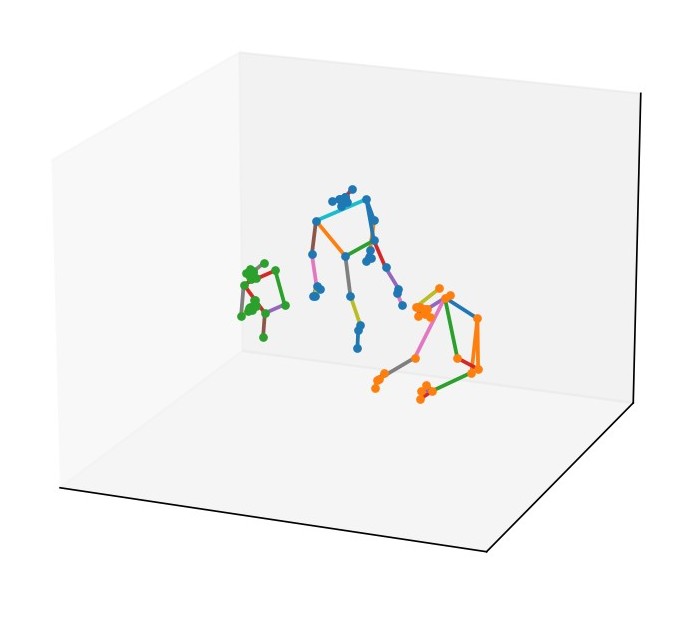}
    \hfill
    \includegraphics[height=3.8cm]{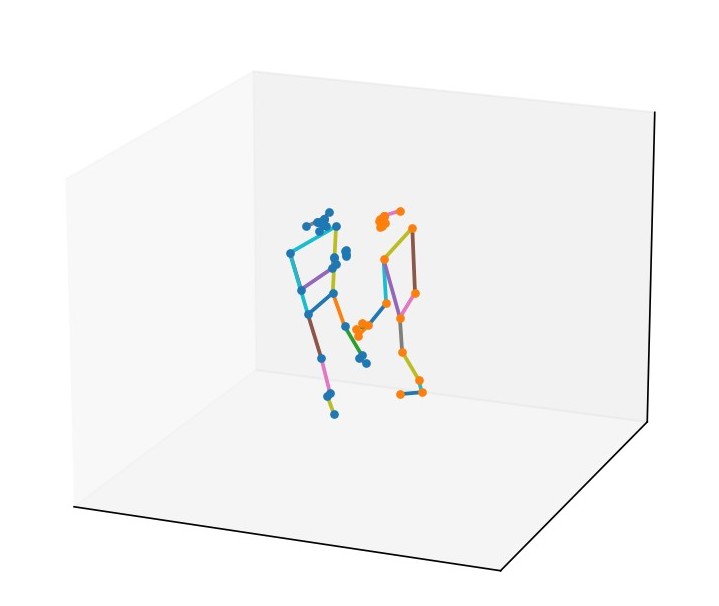}

    \vspace{-0.1cm}
    \caption{Examples of 3D skeleton reconstructions obtained from Camera 1 (\textit{cam1}) and Camera 2 (\textit{cam2}).}
    \label{fig:3d_rec}
\end{figure}

\subsubsection{3D Reconstruction}
For multi-view data, we adopted a geometric 3D reconstruction pipeline that prioritizes true triangulation and uses fusion only to complete missing information. This design follows recent work emphasizing explicit geometric reasoning for handling viewpoint changes and occlusions \cite{liao2024mvgformer, chen20243dsa}.
We first estimated the relative pose between \textit{cam1} and \textit{cam2} using a checkerboard ($10$ $\times$ $7$ inner corners, $24$mm square size), already used for intrinsic calibration. From synchronized frame pairs, 4 inner corners were manually selected and used with \texttt{solvePnP} to estimate the board pose for each camera. The relative transformation between cameras was then derived and conservatively refined by retaining only synchronized frame pairs with low and mutually consistent reprojection error across the two views. In practice, all retained pairs showed reprojection errors below approximately 2 pixels in both cameras.
With temporal alignment fixed, 2D keypoints from both views were triangulated, followed by reprojection-based filtering. A fusion step then completed missing joints, preserving triangulated points when available and relying on single-view estimates only when necessary, as in prior work \cite{liao2024mvgformer}.
Results show that multi-view reconstruction was consistently present but uneven over time, with a moderate number of valid joints (examples depicted in Figure \ref{fig:3d_rec}). The fusion step improved density and continuity, indicating that while triangulation alone was insufficient for complete reconstruction, it provided a reliable geometric backbone. The final output is grounded in true multi-view geometry, with controlled completion only where needed.

\subsubsection{3D Features Set}
Starting from the fused multi-view outputs, we constructed a supervised person-level dataset following the same principles as the 2D pipeline. Each sample corresponds to a temporal window associated with an annotated individual, combining: (i) frame-level 3D outputs from the multi-view/fused pipeline, (ii) identity mapping based on \textit{cam1} tracks, and (iii) stitched poses from \textit{cam1} for alignment.
Dataset construction followed five steps: (1) frame-level 3D outputs were organized into a structured representation per person and frame; (2) identities were assigned, retaining only reliable mappings; (3) a $2$Hz timeline was built for each individual; (4) sliding windows of $2$s, $4$s, and $6$s (stride $0.5$s) were generated and labeled via majority voting; (5) each window was converted into a tabular sample through feature aggregation.
The final dataset includes 10 multi-view videos (20 in total) where all pipeline components were available. After merging, four versions were created: all\_windows, 2s\_only, 2s ML-ready, and a final 2s trainable set. The full $2$s dataset contains 93,906 samples and 23 columns, while quality filtering reduces it to $607$ samples with $10$ numerical features.
The extracted features summarize the quality and coverage of the 3D signal (e.g., number of valid joints, reprojection error, and multi-view/fused ratios), making this a tabular 3D baseline. Notably, the final 3D dataset is almost entirely based on triangulated joints, with negligible single-view contribution, thus representing a fully end-to-end multi-view 3D pipeline grounded in real geometry and consistent identity tracking.



\subsection{Co-Designed Annotation Process}

The entire annotation process was developed through an iterative, co-designed approach involving educator experts, with the explicit goal of creating a \textit{human-centered AI} system that is both usable and useful in their daily practice. The educators' participation ensured that the observation protocol reflects the center's real-world context, is consistent with educators’ needs and workflows, and respects the project's ethical constraints \cite{tang2026disability}. 
The goal was not to produce clinical or diagnostic labels, but rather to define a set of descriptive, observable, and sufficiently stable categories that could be reliably annotated in videos and later used for computational analysis to support and augment educators’ everyday activities.

\subsubsection{Iterative Co-design with Educators}
\begin{figure}[t]
    \centering
    \includegraphics[width=0.83\columnwidth]{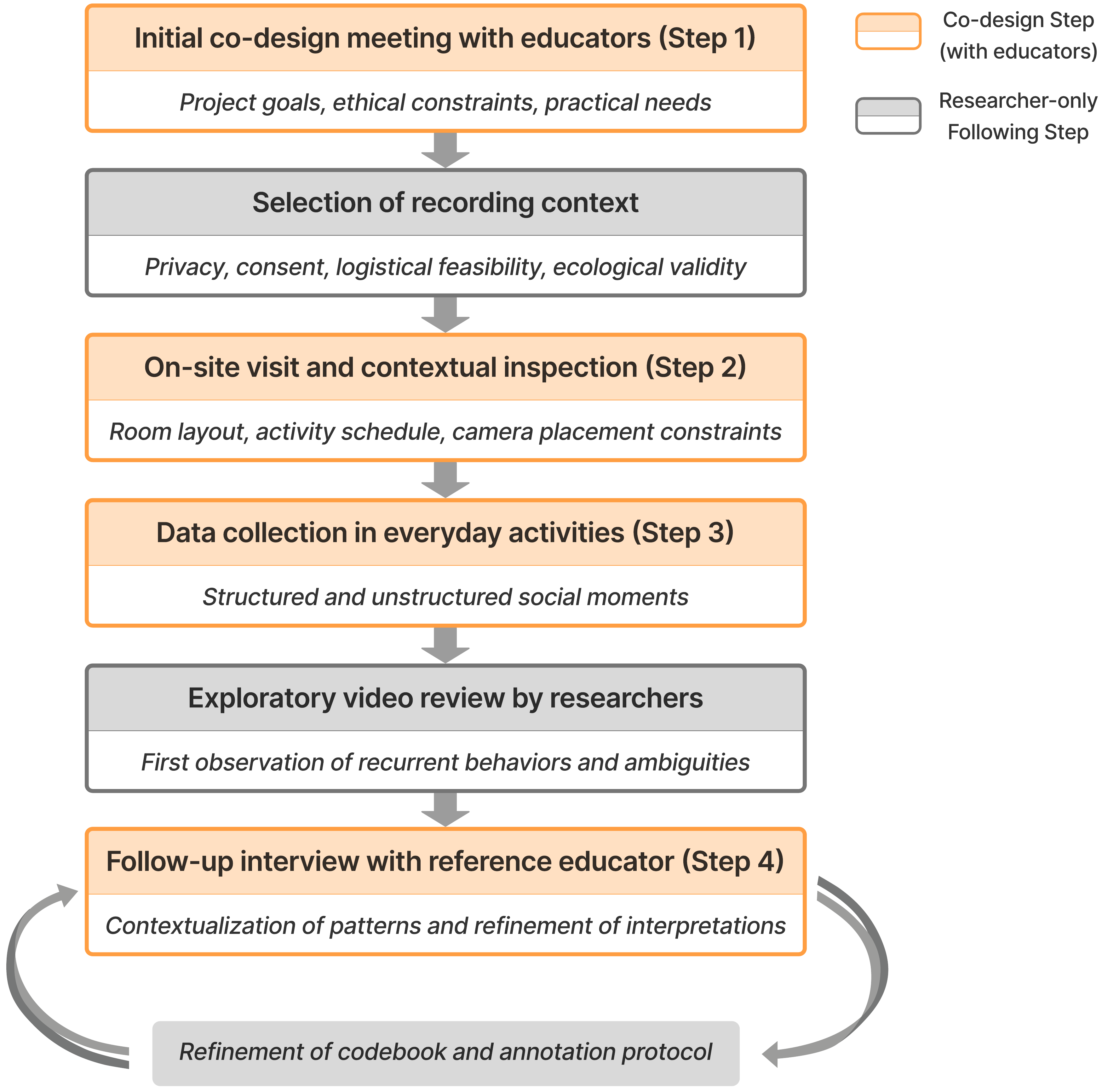}
    \caption{Iterative co-design process adopted to define the annotation protocol and refine the final codebook.}
    \label{fig:codesign_flow}
\end{figure}
The iterative co-design process includes the the following steps (Figure \ref{fig:codesign_flow}).

\textbf{Step 1.} The annotation process was defined through an initial discussion with three educators (two females and one male), aged 45--65, from the Fraternità\&Amicizia daycare center. The initial co-design phase comprised two main interview meetings: the first aimed at distilling the educators' practical needs, and the second focused on proposing and refining a technological solution, which we elaborate on below.
In the first meeting, the project's overall objective, i.e., to analyze social and behavioral dynamics in real-world educational settings using non-invasive signals, was presented, and the practical relevance of this objective to their needs was discussed with the center's staff. This initial discussion allowed us to identify the most appropriate data collection context from an ethical, logistical, and organizational perspective. We transcribed all educators' insights and conducted a data-driven thematic analysis \cite{clarke2017thematic}; the specific codes and themes that emerged are reported in a public GitHub repository%
\footnote{\url{https://github.com/simommmmm/supplementary_materials}}. 
A key need identified was monitoring “relational autonomy” and emotional states to prevent phenomena such as emotional contagion (e.g., the spread of distress among peers). Educators also emphasized the importance of accounting for high individual variability, favoring subject-specific calibration over broad clinical categories.
Based on these needs, in a second meeting we proposed an AI-based observation tool designed to detect micro-dynamics of behavior (e.g., agitation or prolonged isolation) using privacy-preserving signals. The system was envisioned for use during unstructured social moments, such as post-lunch breaks, when participants interact more autonomously.
Educators validated this approach, highlighting three key design principles. First, the system should act as a support tool for real-time alerts, not continuous supervision. Second, it should function as a “tool for reading reality,” flagging observable anomalies (e.g., tension or isolation) without making clinical interpretations. Third, it must allow contextual and personalized interpretation, adapting to each individual’s normative behavior. Also, privacy, consent, and ecological feasibility were identified as critical factors in defining the recording setup and observed activities.

\textbf{Step 2.} Once the data collection environment had been discussed and agreed with the educators, an on-site visit was conducted at the daycare center. During this visit, an educator illustrated to two researchers the available spaces, the usual seating arrangements, the organization of activities during the week, and some recurring characteristics of the group's interactions. From this visit, we learned which locations in the room were most suitable for future camera placement, taking into account both practical constraints (e.g., proximity to power outlets) and observational requirements, such as elevated and angled viewpoints capable of covering the whole room while minimizing blind spots. The visit also clarified the temporal organization of the participants' everyday activities and helped identify the most informative time windows for recording. In particular, we learned that the hours before 9:30 a.m. and after 2:00 p.m. were generally less suitable due to the limited number of participants, and that Fridays were usually less informative because activities often took place outdoor.

\textbf{Step 3.} Data collection was conducted during regular daycare activities led by 4–5 female educators (aged approximately 30–50), who were not involved in the initial co-design phase. The daily schedule included arrival (9:00–9:45), structured morning sessions (9:45–12:40) with activities varying by day (e.g., production, literary analysis, artistic expression, illustrated books), followed by meal preparation (12:40) and lunch (12:45–13:30). Data collection particularly focused on the post-lunch unstructured break (13:30–14:00), where participants engaged more freely in social interactions.

\textbf{Step 4.} After data collection, we conducted a follow-up semi-structured phone interview with the same reference educator to contextualize patterns observed during an initial review of the videos. Instead of showing video excerpts, we discussed a small set of questions derived from recurring situations, such as distinguishing relational tension from normal interaction, interpreting playful physical contact, understanding withdrawn postures (e.g., isolation or bent posture), and identifying behaviors most relevant for monitoring.
The interview was transcribed and used to refine the annotation process. Specifically, this interview clarified some recurring aspects of the relational context: the presence of consolidated emotional bonds between people who have been attending the center for many years; the role of physical contact as a mode of interaction for some participants; the presence, in some cases, of difficulties in regulating negative arousal states, sometimes manifested through more abrupt movements or tensions with other classmates; and, finally, the relevance of rapid visual or bodily orientation cues that can trigger a reaction in others. This information guided the construction of the codebook toward categories that were more consistent with the observed context and more useful for prudent interpretation of the scenes.
The final codebook was not defined \emph{a priori}, but emerged progressively from an iterative process that combined direct observation of the videos, discussion with the educators, and revisions of the annotated categories. 


\subsubsection{Annotation Protocol}

The annotation protocol prioritized directly observable behaviors, avoiding categories based on psychological interpretations or unverifiable inferences, as recommended by educators. Annotations were organized into four tiers: \textit{individual} (A), capturing observable behaviors of a single person; \textit{relational} (B), describing dyadic interactions; \textit{support context} (C), indicating educator interventions; and \textit{tag notes} (D), used to mark ambiguity, occlusions, or visibility issues. This structure separates individual, relational, and contextual information.
All videos were manually annotated using ELAN \cite{wittenburg2006elan} by one researcher, which supports multilevel temporal annotation. Both camera views were synchronized within a single file to enable integrated analysis and reduce occlusion issues. To assess reliability, a second researcher independently annotated 20\% of the dataset, randomly selected at the video level. Agreement was computed over the full annotation structure, covering the four logical tier groups (individual, relational, support context, and tag notes). The comparison was performed on a temporally discretized representation of the doubly annotated videos, using fine-grained temporal bins to compare annotation presence or absence over time. 
Across all tier--time-bin decisions, Cohen's $\kappa$ was 0.804 and raw agreement was 0.840 as reported in Table \ref{tab:annotation_reliability}.

\begin{table}[t]
    \centering
    \footnotesize
    \caption{Inter-annotator reliability across the annotation groups.}
    \label{tab:annotation_reliability}
    \begin{tabular}{lcc}
        \toprule
        \textbf{Annotation group} & \textbf{Raw agreement} & \textbf{Cohen's $\kappa$} \\
        \midrule
        Individual      & 0.833 & 0.313 \\
        Relational      & 0.868 & 0.448 \\
        Support context & 0.891 & 0.579 \\
        Tag notes       & 0.767 & 0.305 \\
        \midrule
        Overall         & 0.840 & 0.804 \\
        \bottomrule
    \end{tabular}
\end{table}

The overall value is computed over the pooled tier--time-bin decisions and therefore should not be interpreted as the arithmetic mean of the tier-specific $\kappa$ values.
Because the double-annotated subset was randomly selected at the video level, it did not guarantee sufficient representation of every individual or relational behavior, particularly the rarest labels. We therefore report reliability at the logical tier-group level rather than per-label Cohen's $\kappa$, which would be unstable or uninformative for sparsely represented behaviors.

The selection of videos for annotation was based on both technical and content-related criteria. From a technical perspective, we evaluated the quality of skeleton extraction across the two camera views using two indicators: the average number of kept poses per frame and the proportion of frames with no valid pose (drop rate). Videos were considered technically robust when both views showed approximately one or more kept poses per frame on average and relatively low drop rates (below $30-35$\%). Conversely, videos were considered technically weak when both cameras exhibited severe pose loss, as in Thursday hour 10:00 (\textit{cam1}: $0.18$ poses/frame, $82.76$\% drop rate; \textit{cam2}: $0.46$, $55.29$\%) and Thursday hour 11:00 (\textit{cam1}: $0.66$, $51.94$\%; \textit{cam2}: $0.20$, $80.58$\%). Frames were considered sufficiently legible when at least one camera provided stable pose coverage.
From a content perspective, we prioritized videos with a sufficient density and diversity of interactions. In practice, we excluded segments where more than $70-80$\% of the time was dominated by routine or already represented activities, and selected those introducing new postural, relational, or interactional patterns.


From an initial set of 20 hourly videos, 10 were selected for annotation: 4 from Monday (H09–H12), 4 from Tuesday (H09, H10, H12, H13), and 2 from Thursday (H09 and a segment of H12 from minute about $38$). Wednesday was entirely excluded due to low interaction density and limited variability, despite acceptable technical quality (\textit{cam1}:$0.97$–$1.76$ poses/frame, \textit{cam2}: $1.01$–$1.77$, drop rates: $6.37$–$32.32$\%).
7 technically robust videos were discarded due to redundancy, while 3 with uneven quality across cameras were retained when at least one view provided sufficient interpretability.
This selection strategy ensured a dataset that was both technically reliable and representative of relevant behaviors.

\subsubsection{Final Codebook \& Labeling}
\begin{table*}[htb!]
\centering
\scriptsize
\renewcommand{\arraystretch}{1.0}
\setlength{\tabcolsep}{2.5pt}
\caption{Final dataset labels with descriptions. Bracketed codes refer to the full annotation codebook available on GitHub.}

\begin{tabularx}{\textwidth}{>{\raggedright\arraybackslash}p{2.6cm} >{\raggedright\arraybackslash}X r r r r r}
\toprule
\textbf{Label} & \textbf{Description} & \textbf{Episodes} & \textbf{Dur. (s)} & \textbf{Cam1} & \textbf{Cam2} & \textbf{3D} \\
\midrule

\roundhl[pink]{Stand} 
& (A1a) Transition from sitting to standing position 
& 141 & 256.34 & 582 & 468 & 176 \\

\roundhl[purple]{Sit} 
& (A1b) Transition from standing to sitting position 
& 139 & 292.89 & 645 & 513 & 75 \\

\roundhl[orange]{ClosedPosture} 
& (A3) Body posture characterized by inward orientation or reduced openness (e.g., hunched or self-contained posture) 
& 87 & 8,335.56 & 14,216 & 10,499 & 236 \\

\roundhl[yellow]{DisengagedRepetitive} 
& (A4 + A5) Disengagement from group activity and/or presence of repetitive movements 
& 28 & 2,070.11 & 1,929 & 100 & 120 \\

\midrule
\multicolumn{2}{l}{\textbf{Individual benchmark total}}
& & 
& \textbf{17,372}
& \textbf{11,580}
& \textbf{607} \\
\midrule

\roundhl[blue]{SocialOrientation} 
& (B1 + B4) Physical proximity and mutual orientation toward another person 
& 41 & 674.22 
& 885
& 164 
& 0 \\

\roundhl[green]{AffiliativeContact} 
& (B2) Positive social physical contact (e.g., supportive or friendly touch) 
& 62 & 2,322.22 
& 724
& 325 
& 3$^{*}$ \\

\midrule
\multicolumn{2}{l}{\textbf{Relational retained windows}}
& &
& \textbf{1,609}
& \textbf{489}
& \textbf{3$^{*}$} \\
\midrule

\multicolumn{2}{l}{\textbf{Annotated events total}}
& \textbf{498}
& \textbf{13,951.34}
& \multicolumn{3}{c}{\textit{Task-specific window totals reported above}} \\

\bottomrule
\end{tabularx}

\vspace{0.35em}

\begin{minipage}{\textwidth}
{\footnotesize
Episodes and Dur. report the number and aggregate duration of the original annotated events. Cam1, Cam2, and 3D report derived 2\,s windows: person-level for the individual task and pair-level for the relational task. Windows use a 0.5\,s stride and therefore overlap. $^{*}$Only three usable 3D relational windows remained, all belonging to AffiliativeContact; therefore, no binary 3D relational benchmark could be constructed. Person- and pair-level counts are reported separately and are not summed.
}

\end{minipage}

\label{tab:final_class_distribution}
\end{table*}

The codebook was structured into four macro-areas: \textit{individual} behaviors, \textit{relational} interactions, \textit{support context} (educator interventions), and \textit{tag notes} (ambiguity or visibility issues). Individual labels describe observable behaviors of a single subject (e.g., posture, transitions, repetitive actions), while relational labels capture dyadic configurations such as proximity, orientation, and contact.
To ensure consistency, all labels were defined using observable criteria and, when needed, temporal thresholds: ClosedPosture (A3) and DisengagedBehaviour (A4) required at least $5$s; MutualOrientation (B4) at least $2$s; Proximity (B1) at least $5$s and non-accidental; RepetitiveMovement (A5) required either three repetitions within $10$s or persistence for at least $5$s.
A standalone summary of the complete annotation codebook is reported in Appendix C. 

The annotation distribution shows strong class imbalance, with frequent classes such as Stand (A1a), Sit (A1b), and especially ClosedPosture (A3), and rarer ones including DisengagedBehaviour (A4), MutualOrientation (B4), and PossibleTension (B5). To address this, labels were aggregated for the supervised setting: Stand, Sit, and ClosedPosture were retained, while DisengagedBehaviour and RepetitiveMovement were merged into DisengagedRepetitive. This aggregation is a modeling compromise motivated by label sparsity, rather than evidence that repetitive movement and disengagement represent the same construct. Predictions of the merged class cannot distinguish its constituent behaviors and should not be interpreted as establishing disengagement or any particular affective state. Similarly, ClosedPosture denotes a bodily configuration and does not, by itself, establish discomfort or distress. These distinctions constrain the interpretation and potential downstream use of benchmark predictions.
We have also aggregated relational aspects into: SocialOrientation and AffiliativeContact.
The final class distribution is reported in Table~\ref{tab:final_class_distribution}. 
Both the dataset (including skeleton-based features) and the labels (with the corresponding codebook) are included in the GitHub repository.

\section{Body Behavior Recognition Modeling}
This section presents baseline recognition experiments on two complementary tasks derived from the annotated data: an individual-level four-class task and a relational pair-level binary task. Both rely on observable, skeleton-derived cues and provide controlled and interpretable benchmarks consistent with the project's ethical constraints, avoiding biometrics, face recognition, and voice analysis.

\subsection{Problem Formulation}
In this work, we formulate the \textit{individual task} as body behavior recognition of single individuals, i.e., 4-class supervised classification of observable behaviors based on skeletal features extracted from person-level tracking over time. Each sample corresponds to a person within a temporal window and consists of a feature vector describing their behavior, along with the associated label. Labels are derived from the annotation codebook and subsequently reduced to a smaller set to address sparsity and ensure statistical stability.
For the \textit{relational task}, each sample corresponds to a pair of participants observed within the same temporal window, rather than to a single tracked individual. The target classes are SocialOrientation, obtained by aggregating Proximity (B1) and MutualOrientation (B4), and AffiliativeContact (B2). Only windows for which both members of the pair could be reliably associated with the available pose information were retained. This formulation therefore introduces an additional observability requirement with respect to the individual-level task, since both participants must be simultaneously represented for a relational sample to be usable.

The adopted formulation falls within the broader area of \textit{skeleton-based action recognition}, but with some relevant differences from the most common settings. First, the observed context is ecological and uncontrolled, with high postural variability, occlusions, seated subjects, and spontaneous interactions. Second, the focus is not on athletic actions or standardized gestures, but on socially relevant body behaviors, often extended in time and characterized by less sharply defined temporal boundaries. Finally, the modeling stage was built on interpretable features derived from 2D poses and tracking, to prioritize transparency and model readability in the initial phase. This choice differs from more complex approaches based on deep backbones for skeleton action recognition or on pose-derived volumetric representations, which nonetheless remain important reference for future work \cite{duan2022poseconv3d}.

\subsection{Dataset Definition}
The supervised dataset integrates three components: manual temporal annotations, person-level tracking from skeleton extraction, and features aggregated over temporal windows. Each sample corresponds to a temporal window associated with a tracked and annotated individual.
For each reliably matched subject, a timeline was discretized at $2$ Hz, balancing temporal resolution and robustness. Sliding windows of $2$s, $4$s, and $6$s were then applied with a stride of $0.5$s. 
Because consecutive windows overlap substantially, the resulting window-level samples are temporally dependent and should not be interpreted as independent behavioral observations. In particular, the main 2\,s configuration with a 0.5\,s stride results in 75\% overlap between consecutive windows. Table~\ref{tab:final_class_distribution} therefore distinguishes the original annotated behavioral episodes and their durations from the derived window counts used for modeling.
Each window was assigned a label based on majority voting, requiring a single class to cover more than $50$\% of the samples. Ambiguous windows were discarded from the final dataset.
From an operational perspective, the baseline included only instances where the association between annotated identity and stitched track was reliable. Specifically, we retained only participant tracks with clear and unambiguous identity matching, as verified through manual inspection, and deemed suitable for downstream modeling.
Accordingly, educators, ambiguous cases, and unreliable tracks were excluded. 
The final classes used in the baseline dataset configuration were: \roundhl[purple]{Sit}, \roundhl[pink]{Stand},  \roundhl[orange]{ClosedPosture}, and \roundhl[yellow]{DisengagedRepetitive}. 
These classes include two general atomic actions (Sit and Stand), used as reference action-recognition categories, and two more domain-specific behaviors (ClosedPosture and DisengagedRepetitive). The main benchmark should therefore be interpreted as a person-level body behavior recognition task rather than as a complete evaluation of socially aware interaction recognition.
The features for each window were aggregated into an interpretable representation capturing tracking continuity, pose quality, bounding-box dynamics, and coarse body measures from the 2D skeleton. Specifically, features include: number of tracks, number of valid pose frames, temporal coverage, pose quality statistics, bounding-box center (mean and variability), size and area, velocity and total displacement, and body measures such as shoulder width, hip width, and torso height.
This design prioritizes simple and interpretable features as a baseline before exploring more complex representations.
In the baseline version of the dataset, separate sets were created for \textit{cam1} and \textit{cam2}. 
Restricting the analysis to $2s$ windows only, the number of samples is $17,372$ for \textit{cam1} and $11,580$ for \textit{cam2}. The class distribution is strongly imbalanced, with a clear predominance of the class \roundhl[orange]{ClosedPosture}, which dominates both views, as reported in Table \ref{tab:final_class_distribution}. This imbalance directly reflects the nature of the observed context and represents one of the key aspects to consider when interpreting the baseline results.

For the relational task, 2\,s pair-level windows were generated independently for the two camera views, yielding 5,961 candidate windows per camera. After identity mapping, 5,709 windows remained in which both members of the annotated pair could be mapped, comprising 1,049 \roundhl[blue]{SocialOrientation} and 4,660 \roundhl[green]{AffiliativeContact} instances. Given this substantial class imbalance, AffiliativeContact was downsampled to match the 1,049 SocialOrientation instances, resulting in a balanced benchmark of 2,098 windows. The final set contains 1,609 windows from \textit{cam1} and 489 from \textit{cam2}; specifically, \textit{cam1} contributes 885 \roundhl[blue]{SocialOrientation} and 724 \roundhl[green]{AffiliativeContact} windows, while \textit{cam2} contributes 164 and 325, respectively.

To support privacy-preserving reuse, the dataset is released as a derived-feature benchmark rather than as a raw audiovisual dataset. The public release does not include raw video, audio, or directly identifying information. Instead, it provides the annotation codebook, dataset documentation, data dictionaries, and the derived tabular datasets used for the \textit{cam1}, \textit{cam2}, and multi-view 3D benchmarks. 

\subsection{Baseline Results}
We evaluate the two recognition tasks as parallel but distinct benchmarks: an individual-level four-class task and a relational pair-level binary task. The monocular individual baselines and the relational benchmark use video-level grouped partitions to reduce temporal and recording-specific leakage across training and test data, whereas the individual 3D SVM branch uses the fixed holdout protocol described below.

\begin{table}[htb!]
    \centering
    \scriptsize
    \caption{Baseline results from \textit{cam1} for SVM and PoseC3D.}
    \setlength{\tabcolsep}{2pt}
    \renewcommand{\arraystretch}{1.18}
    \resizebox{\columnwidth}{!}{%
    \begin{tabular}{lccccc}
    \toprule
    \textbf{Setting} & \textbf{Model*} & \textbf{Acc.} & \textbf{Bal. Acc. /} & \textbf{Macro-F1 /} & \textbf{W-F1 /} \\
     &  &  & \textbf{Mean Acc.} & \textbf{Top-1} & \textbf{Notes} \\
    \midrule
    all\_windows & SVM & 0.61 & 0.24 & 0.27 & \textbf{0.70} \\
    2s\_only & SVM & \textbf{0.69} & \textbf{0.30} & \textbf{0.31} & \textbf{0.70} \\
    2s\_only + under 0.30 & SVM & 0.58 & 0.28 & 0.30 & 0.62 \\
    2s\_only + under 0.50 & SVM & 0.58 & 0.28 & 0.29 & 0.62 \\
    \addlinespace[2pt]
    \midrule
    2s\_only & PoseC3D & -- & 0.33 & 0.94 & highly unstable \\
    all\_windows pose-avail. & PoseC3D & -- & 0.33 & 0.98 & slightly more promising \\
    \bottomrule
    \end{tabular}%
    }
    \vspace{2pt}
    {\footnotesize *SVM reports accuracy, balanced accuracy, macro-F1, and weighted-F1; PoseC3D reports best mean class accuracy and top-1 accuracy.}
    \label{tab:cam1_baseline_results}
\end{table}


\subsubsection{Individual-level recognition}
The task was initially formulated as a multiclass classification over sliding temporal windows, using an SVM as an interpretable baseline. While more advanced models from skeleton-based recognition literature were considered \cite{multimediate2023,conflab2022}, we adopted a simple setup to establish a solid reference point.
Evaluation was first conducted on the \textit{cam1} view, treating it as the reference configuration. Alongside the SVM baseline, we also performed an initial exploratory experiment using PoseC3D.
For the SVM baseline, we evaluated four configurations on the \textit{cam1} view using the following classes: \roundhl[purple]{Sit}, \roundhl[pink]{Stand},  \roundhl[orange]{ClosedPosture}, and \roundhl[yellow]{DisengagedRepetitive}.
All experiments used GroupKFold (grouped by \texttt{video\_id}),
StandardScaler, and an SVC classifier with
\texttt{class\_weight='balanced'}, while considering multiple
window lengths (2s, 4s, and 6s).
Grouping by \texttt{video\_id} prevents windows from the same recording from appearing in both training and test folds. However, it does not enforce participant-disjoint splits, since the same participant may occur in different videos. The reported results should therefore be interpreted as cross-video generalization rather than generalization to unseen participants. 
Participant-level label distributions are reported in Appendix B  
to make potential participant--label dependencies transparent.


Hyperparameters were selected via GridSearchCV, and performance was evaluated using macro-F1 as the primary metric. Given the strong class imbalance of the benchmark, balanced accuracy and macro-F1 are used as the primary metrics for model interpretation, while raw accuracy and weighted-F1 are reported as complementary measures. To contextualize model performance, we additionally evaluated two baselines: a majority-class baseline, which always predicts the most frequent class in the training data, and a class-prior baseline, which samples predictions according to the empirical class distribution of the training data. For \textit{cam1} and \textit{cam2}, these baselines were evaluated using the same video-level GroupKFold partitions as the SVM models. For the 3D branch, they were evaluated on the same fixed holdout split used for the corresponding SVM baseline.

\begin{table}[t]
\centering
\scriptsize
\setlength{\tabcolsep}{2pt}
\renewcommand{\arraystretch}{1.0}

\caption{Diagnostic comparison of SVM and dummy baselines for the individual-level recognition task.}
\label{tab:dummy_baselines}

\resizebox{0.7\columnwidth}{!}{%
\begin{tabular}{llcccc}
\toprule
\textbf{Setting} & \textbf{Model} & \textbf{Acc.} & \textbf{Bal. Acc.} & \textbf{Macro-F1} & \textbf{W-F1} \\
\midrule
cam1 2s & Majority-class & 0.796 & 0.267 & 0.234 & 0.720 \\
        & Class-prior    & 0.640 & 0.249 & 0.215 & 0.643 \\
        & SVM            & 0.691 & 0.302 & 0.314 & 0.696 \\
\midrule
cam2 2s & Majority-class & 0.829 & 0.267 & 0.240 & 0.759 \\
        & Class-prior    & 0.745 & 0.264 & 0.235 & 0.725 \\
        & SVM            & 0.726 & 0.352 & 0.320 & 0.788 \\
\midrule
3D      & Majority-class & 0.388 & 0.250 & 0.140 & 0.217 \\
        & Class-prior    & 0.322 & 0.263 & 0.261 & 0.316 \\
        & SVM            & 0.520 & 0.380 & 0.342 & 0.443 \\
\bottomrule
\end{tabular}%
}
\end{table}

Results are reported in Table~\ref{tab:dummy_baselines}.
Among the tested \textit{cam1} configurations, the 2s-only dataset provided the strongest standard SVM baseline, reaching a balanced accuracy of 0.302 and a macro-F1 of 0.314 (accuracy 0.691). Restricting the dataset to 2s windows improved macro-F1 from 0.27 for the all-windows configuration to 0.31, suggesting that a temporally consistent representation is more effective for this baseline than combining multiple window durations.
Table~\ref{tab:dummy_baselines} highlights why raw accuracy should be interpreted cautiously under the observed class imbalance. For \textit{cam1} and \textit{cam2}, the majority-class baselines achieve higher accuracy than the corresponding SVMs, but substantially lower balanced accuracy and macro-F1. For the 3D branch, the SVM outperforms both dummy baselines across all reported metrics. We therefore consider balanced accuracy and macro-F1 the most informative metrics for benchmark interpretation.
The two experiments with undersampling of \roundhl[orange]{ClosedPosture} (keep\_ratio = 0.30 and 0.50) did not improve the results, but instead worsened the global metrics, confirming that the main limitation of the baseline does not depend only on class imbalance, but also on the video-specific structure of the dataset.
Per-class precision, recall, and F1 scores, together with the corresponding confusion matrices, are reported in Appendix B. 
Alongside the SVM baseline, we conducted an exploratory evaluation using PoseC3D on the \textit{cam1} view. Experiments were run on both the $2$s-only and all-windows (pose-available) datasets. While training was successful in both cases, validation results were highly unstable. In particular, top-1 accuracy was very high (up to $0.94$ for $2$s-only and $0.98$ for all-windows), but mean class accuracy remained low (around $0.33$). 
This discrepancy indicates that the model is largely driven by the dominant class and does not yet achieve reliable multiclass discrimination.

Overall, the results on \textit{cam1} indicate that the most solid and interpretable baseline, at the current stage, remains SVM + $2$s\_only standard. 
A similar pattern was observed for \textit{cam2}, where top-1 accuracy reached 0.96 while mean class accuracy remained fixed at approximately 0.33, further indicating that high top-1 scores in the monocular PoseC3D experiments should not be interpreted as robust multiclass recognition.
The PoseC3D experiments are promising as a future direction, but still too unstable to be considered a consolidated baseline.
The discrepancy between overall and class-balanced performance illustrates why aggregate accuracy is insufficient for evaluating recognition in this setting. A model can achieve high accuracy while providing limited discrimination for less frequent behaviors. Evaluation should therefore make performance on each recognition target visible, while recognizing that neither class-balanced metrics nor benchmark accuracy establish usefulness in practice.

\subsubsection{Relational recognition}
For the relational track, we evaluated an SVM on the balanced 2\,s pair-level benchmark comprising 2,098 windows, with 1,049 instances for each relational category. Evaluation followed a five-fold video-level grouped protocol, consistently with the monocular individual-level baselines. The relational SVM achieved an accuracy and balanced accuracy of 0.367, a macro-F1 of 0.305, and a weighted-F1 of 0.305.
Performance differed substantially across the two relational categories. \roundhl[blue]{SocialOrientation} achieved a precision of 0.417, recall of 0.665, and F1-score of 0.512, whereas \roundhl[green]{AffiliativeContact} achieved a precision of 0.170, recall of 0.069, and F1-score of 0.098. The corresponding confusion matrix shows that 698 of 1,049 \roundhl[blue]{SocialOrientation} windows were correctly classified, compared with only 72 of 1,049 \roundhl[green]{AffiliativeContact} windows; 977 AffiliativeContact windows were instead classified as SocialOrientation. Despite balancing the two classes before training, the model therefore exhibits limited cross-video discrimination between the relational categories. This suggests that the difficulty of the relational task cannot be attributed to class frequency alone, but is also associated with recording-specific variability and the requirement for reliable simultaneous observation of both participants.

\subsection{Multi-view Results}

\begin{table}[tb!]
    \centering
    \footnotesize
    \caption{Best individual-level baseline results across \textit{cam1}, \textit{cam2}, and multi-view 3D settings.}
    \label{tab:multiview_results}

    \setlength{\tabcolsep}{2.2pt}
    \renewcommand{\arraystretch}{1.08}

    \begin{tabularx}{\columnwidth}{@{}l X c c c c@{}}
    \toprule
    \textbf{Setting} &
    \textbf{Configuration} &
    \textbf{Acc./} &
    \textbf{Bal./} &
    \textbf{Macro-} &
    \textbf{W-F1/} \\
    &
    &
    \textbf{Top-1} &
    \textbf{Mean} &
    \textbf{F1} &
    \textbf{Notes} \\
    \midrule

    \textit{cam1} &
    SVM, 2s-only &
    0.69 & 0.30 & 0.31 & 0.70 \\

    \textit{cam1} &
    PoseC3D, all-windows &
    \textbf{0.98} & \textbf{0.33} & -- & unstable \\

    \midrule

    \textit{cam2} &
    SVM, 2s-only &
    0.73 & \textbf{0.35} & 0.32 & 0.79 \\

    \textit{cam2} &
    PoseC3D, all-windows &
    \textbf{0.96} & 0.33 & -- & flat val. \\

    \midrule

    3D &
    SVM, 2s trainable &
    0.52 & 0.38 & 0.34 & 0.44 \\

    3D &
    PoseC3D export &
    \textbf{0.68} & \textbf{0.59} & -- & 249 samples \\

    \bottomrule
    \end{tabularx}

    \vspace{0.7ex}

    {\footnotesize
    SVM reports accuracy, balanced accuracy, macro-F1, and weighted-F1; PoseC3D reports top-1 and mean class accuracy.
    }
\end{table}
\paragraph{Individual-level recognition}
To assess the impact of multi-view data, we compared three settings: monocular \textit{cam1}, monocular \textit{cam2}, and the multi-view 3D dataset obtained from the triangulation and fusion pipeline. Table~\ref{tab:multiview_results} summarizes the best results obtained with both SVM and PoseC3D across the three settings.
For the monocular views, the best SVM configuration was 2s\_only in both cases. \textit{Cam2} yielded the strongest monocular SVM result, with accuracy $0.73$, balanced accuracy $0.35$, and macro-F1 $0.32$, slightly outperforming \textit{cam1}. In both views, 2s\_only was preferable to all\_windows, while undersampling did not lead to stable improvements.
The multi-view 3D branch resulted in a smaller but higher-quality dataset ($607$ samples, $10$ features). Its SVM baseline achieved accuracy $0.52$, balanced accuracy $0.38$, and macro-F1 $0.34$, yielding the best balanced accuracy and macro-F1 among the SVM settings.

The PoseC3D-compatible export derived from the 3D branch reached a top-1 accuracy of 0.68 and a mean class accuracy of 0.59 on 249 exported samples. Because this experiment uses a smaller derived subset and a different evaluation setup from the tabular SVM branches, it is reported as exploratory and should not be interpreted as a direct performance comparison.
To isolate the contribution of the different views from differences in dataset size and filtering, we additionally performed a controlled same-window comparison. 
To assess whether the tabular 3D branch captures behaviorally relevant 3D information beyond reconstruction quality, we performed an additional feature ablation by deriving kinematic descriptors from the reconstructed 3D poses on the subset of windows for which these quantities could be reliably computed. The analysis was restricted to the 288 windows for which valid kinematic descriptors could be computed, and the same split was used for all three feature configurations. Quality features describe reconstruction coverage and reliability, whereas kinematic features capture body geometry and motion, including body dimensions and compactness, joint and body-center velocities, motion energy, and depth variation.

\begin{table}[t]
\centering
\footnotesize
\setlength{\tabcolsep}{3pt}
\renewcommand{\arraystretch}{1.0}
\caption{Feature ablation on the common 288-window 3D subset. All configurations use the same evaluation split.}
\label{tab:3d_feature_ablation}
\begin{tabular}{lrrrr}
\toprule
\textbf{Features} & \textbf{Acc.} & \textbf{Bal. Acc.} & \textbf{Macro-F1} & \textbf{W-F1} \\
\midrule
Quality only   & 0.514 & 0.464 & 0.463 & 0.527 \\
Kinematic only & 0.694 & 0.699 & 0.671 & 0.710 \\
Combined       & 0.778 & 0.768 & 0.747 & 0.783 \\
\bottomrule
\end{tabular}
\end{table}

As shown in Table~\ref{tab:3d_feature_ablation}, kinematic features substantially outperform quality-only descriptors, while the combined representation achieves the strongest results. This indicates that the 3D branch contains discriminative information related to body geometry and motion rather than merely encoding reconstruction reliability. Because this ablation is restricted to the 288 windows with valid kinematic features, its scores should not be directly compared with the results obtained on the full 607-window 3D dataset.
The analysis was restricted to 259 person-level windows shared across \textit{cam1}, \textit{cam2}, and the 3D branch, spanning eight videos. All representations were therefore evaluated on the same samples using video-level GroupKFold. Four configurations were compared: \textit{cam1}-only, \textit{cam2}-only, concatenated \textit{cam1}+\textit{cam2}, and 3D-only features.

\begin{table}[t]
\centering
\footnotesize
\setlength{\tabcolsep}{3pt}
\renewcommand{\arraystretch}{1.0}
\caption{Same-window comparison across monocular, feature-level fusion, and 3D representations.} 
\label{tab:same_window_multiview}
\begin{tabular}{lrrrr}
\toprule
\textbf{Representation} & \textbf{Acc.} & \textbf{Bal. Acc.} & \textbf{Macro-F1} & \textbf{W-F1} \\
\midrule
cam1-only       & 0.279 & 0.229 & 0.138 & 0.232 \\
cam2-only       & 0.363 & 0.309 & 0.220 & 0.353 \\
cam1+cam2       & 0.363 & 0.292 & 0.226 & 0.354 \\
3D-only         & 0.174 & 0.162 & 0.130 & 0.196 \\
\bottomrule
\end{tabular}
\end{table}
On this controlled subset, \textit{cam2} provides the strongest monocular representation. Simple feature concatenation of \textit{cam1} and \textit{cam2} yields only a small increase in macro-F1 (0.226 versus 0.220) and weighted-F1, while balanced accuracy decreases from 0.309 to 0.292. The 3D representation is the weakest in this controlled comparison. These results therefore provide only limited evidence of multi-view complementarity under simple feature-level fusion.

The branch-level and same-window comparisons (Table \ref{tab:same_window_multiview}) provide complementary perspectives. While the full 3D branch achieves competitive balanced accuracy on its independently filtered dataset, this advantage does not persist when all representations are restricted to the same person-window samples. Under the controlled comparison, \textit{cam2} performs best overall, whereas simple \textit{cam1}+\textit{cam2} concatenation provides only a modest gain in macro-F1. We therefore interpret the current multi-view results cautiously: the additional viewpoint is useful for geometric reconstruction and increases observational coverage, but the present tabular fusion baseline does not demonstrate a substantial classification advantage over the strongest monocular view.
A dedicated stratification by occlusion or pose coverage remains necessary to quantify more directly when one view compensates for failures of the other.
The difference between branch-level and matched-window results highlights the need to separate representation quality from sample selection. Comparisons between independently filtered branches cannot isolate the benefit of an additional viewpoint. The matched-window analysis addresses this issue for retained observations, but does not establish whether filtering excludes particular participants or behaviors.

\paragraph{Relational recognition}
The multi-view analysis reveals a substantially different limitation for the relational task. Among the 5,709 pair-level windows retained after identity mapping, at least one participant had available 3D information in 792 windows, whereas both members of the pair were simultaneously represented in only 37 windows. After applying the fused multi-view representation, only 30 windows retained usable information for both participants, all belonging to AffiliativeContact; no SocialOrientation window satisfied this requirement. In the balanced relational benchmark, this number was further reduced to only three windows. Consequently, a meaningful binary 3D relational classification benchmark could not be constructed.
This result highlights an observability constraint that is specific to relational recognition. While individual behavior recognition requires a reliable representation of one person, relational recognition requires both members of an interacting pair to be simultaneously visible, tracked, associated across views, and represented by the reconstruction pipeline. Multi-view sensing therefore increases the available geometric information but does not, in the present dataset, provide sufficient joint pair coverage to support a 3D relational recognition task.


\section{Discussion \& Outlook}
The contributions of this work are best understood in light of the original motivation behind the creation of \textit{SocialMirror}, which emerged from the design of an assistive, human-centered system aimed at supporting educators in monitoring social and behavioral dynamics in daycare environments for individuals with IDD. In this context, educators highlighted the difficulty of continuously observing multiple individuals and identifying subtle, non-verbal cues—particularly during unstructured activities—revealing a clear mismatch between real-world needs and the capabilities of existing datasets and models. As a result, \textit{SocialMirror} was developed as a foundational resource grounded in ecological validity, capturing naturalistic interactions and socially meaningful behaviors in real care settings. 

The contribution of SocialMirror extends beyond representation within a benchmark \cite{huang2026inclusive, li2023mmasd, shah2026axon, cibrian2025limitations}. 
The study makes visible several decisions that shape the applicability of affective computing in a disability context: how recognition targets are defined, how technical filtering determines which observations remain available, and how evaluation supports or limits claims about generalization. 
The resulting dataset provides a basis for investigating these decisions. Its inclusion of participants with IDD should not itself be interpreted as evidence that the resulting models are equitable, generalizable, or useful in care practice.
A key contribution of this work lies 
also in the process through which the dataset was built. The annotation codebook was developed in close collaboration with educators, ensuring that labels reflect observable and meaningful behaviors grounded in real care practices. This aligns with HCAI principles, which emphasize participatory design and the inclusion of domain expertise in system development \cite{shneiderman2022human, panigutti2023co, almufareh2024conceptual}. 
Our findings reinforce that purely data-driven approaches are insufficient in such contexts without strong domain grounding, particularly when dealing with subtle non-verbal behaviors and vulnerable populations.

Then, the baseline experiments highlight the intrinsic difficulty of body behavior recognition in real-world daycare settings. While the SVM model achieved moderate performance, the diagnostic baselines confirm that raw accuracy is strongly affected by class imbalance, making balanced accuracy and macro-F1 more informative for benchmark interpretation. 
Similarly, the exploratory PoseC3D experiments achieved very high top-1 accuracy but low mean class accuracy, indicating a strong bias toward dominant classes rather than robust multiclass discrimination. These findings are consistent with prior work showing that real-world, socially complex environments remain challenging for current behavior recognition models \cite{liu2025systematic, duan2022revisiting}. 
The moderate performance of baseline models reflects both dataset complexity and current methodological limitations. Future work should explore more expressive temporal models, improved multi-view fusion, and more robust relational modeling under joint-observability constraints. At the same time, maintaining privacy-preserving and ethically grounded approaches remains essential, particularly in sensitive care environments \cite{almufareh2024conceptual}.



\section{Lessons for Inclusive Affective Computing}

SocialMirror connects educator-informed annotation with individual and relational recognition experiments in a naturalistic daycare setting. Together, these findings yield four methodological lessons for developing affective computing systems that accommodate diverse bodily expressions and the social contexts in which they occur.

\textbf{Lesson 1 — Define recognition targets through situated knowledge of everyday interactions.}
Educator interviews revealed that physical contact, withdrawn postures, and changes in orientation need to be understood in relation to established relationships and individual communication practices. These insights informed a codebook that distinguishes individual behaviors, dyadic interactions, educator support, and interpretive uncertainty. The contribution of this process was therefore more specific than selecting relevant behaviors: it translated contextual knowledge into an explicit annotation structure.
For inclusive affective computing, this suggests involving people familiar with the interaction setting when defining recognition targets and documenting the observable criteria, contextual cues, and ambiguities associated with each label. Such a structure provides an intermediate layer between bodily signals and their possible affective significance, supporting subsequent interpretation by practitioners.
This emphasis on situated relationships is consistent with Bircanin et al.~\cite{bircanin2025co}, who examine how meaningful co-design with adults with intellectual disabilities develops through engagement with their everyday contexts. SocialMirror illustrates how contextual knowledge can also inform the construction of computational labels through collaboration with educators.

\textbf{Lesson 2 — Inclusive modeling must account for diversity within the target population.}
SocialMirror shows why including adults with IDD in a benchmark must be accompanied by attention to differences in their behavioral repertoires. The participant-level analysis reveals substantial variation in who contributes to each recognition target: for example, \textit{DisengagedRepetitive} is represented by three participants in cam1 and only one in cam2. Educators also emphasized that bodily behaviors acquire meaning in relation to individual histories and characteristic ways of communicating. Together, these findings motivate treating individual variability as a consideration in model development, rather than assuming that a shared disability category defines a homogeneous recognition problem.
For inclusive affective computing, this means documenting participant--behavior distributions and using them to guide data collection, model design, and evaluation. Population-level models can provide a shared starting point, while individual calibration and adaptation offer research directions for accommodating distinctive behavioral patterns. Comparisons between these approaches should distinguish performance for familiar participants from transfer to previously unseen individuals.
This perspective is consistent with Trewin's emphasis on the diversity of disability-related characteristics~\cite{trewin2018ai} and with evidence that participant overlap affects reported affect-recognition performance~\cite{moreno2024sneaky}. SocialMirror contributes a concrete basis for investigating this issue in body-behavior recognition: a benchmark should characterize both which behaviors are represented and whose ways of expressing them the models encounter.

\textbf{Lesson 3 — Design sensing around the interaction that must be observed.}
The relational analysis reveals a distinction between observing a person and obtaining a usable representation of an interaction. Among 5,709 pair-level windows retained after identity mapping, 792 had 3D information for at least one participant, whereas only 37 had information for both. After fusion, 30 windows remained usable for both participants, all belonging to \textit{AffiliativeContact}. Thus, individual pose availability does not adequately describe the sensing requirements of a relational task.
For systems intended to support interpretation of social and affective dynamics, camera placement, tracking, and filtering should therefore be assessed in terms of simultaneous coverage of interacting people. Pair-level retention should be reported by behavior category, alongside individual pose quality and classification performance. This makes it possible to identify whether a recognition target is lost during sensing or remains difficult after a usable representation has been obtained.
ConfLab~\cite{conflab2022} similarly treats camera viewpoints, body-keypoint visibility, and interaction annotations as central components of naturalistic social-behavior analysis. SocialMirror adds a concrete lesson for care settings: joint observability is a task-specific design criterion that must be measured explicitly trading it off with discomfort of participants feeling monitored by multiple cameras.

\textbf{Lesson 4 — Evaluate progress through coverage of the behavioral repertoire.}
The diagnostic baselines show that the choice of evaluation metric changes which model appears preferable. For \textit{cam1}, the majority-class baseline achieves higher accuracy than the SVM (0.796 versus 0.691), while the SVM achieves higher macro-F1 (0.314 versus 0.234). Moreover, undersampling the dominant class does not improve the individual baseline, and balancing the relational classes does not yield reliable discrimination. These findings indicate that equalizing class counts alone is insufficient to address the recognition problem.
Inclusive evaluation should therefore combine class-balanced metrics with per-behavior precision and recall, episode counts, and participant coverage. This combination distinguishes improvements in recognizing common behaviors from improvements that broaden the range of behaviors a system can recognize, and helps direct further annotation and modeling effort.
Trewin~\cite{trewin2018ai} highlights the challenges that highly diverse disability-related characteristics pose for conventional approaches to AI fairness. Our results give this concern a concrete evaluation implication: progress should be assessed through the distribution of recognition performance across the targeted behaviors, alongside aggregate accuracy.


Together, these lessons position inclusion as a methodological responsibility spanning construct definition, participation, sensing, modeling, and evaluation. SocialMirror provides empirical observations that motivate these adjustments, while leaving their effectiveness and broader applicability open to validation.


\section{Limitations \& Conclusions}
A limitation concerns participant-level generalization: because the evaluation is grouped by video rather than by participant, the same individual may appear in both training and test folds. The benchmark therefore does not establish performance on unseen participants, and participant-specific or context-specific patterns may partly contribute to the reported results.
Participant-disjoint evaluation remains an important direction for future work as larger and more balanced longitudinal datasets become available.
Annotation reliability was assessed on 20\% of the dataset, while the remaining annotations were produced by a single annotator. Moreover, the randomly selected double-annotated subset does not ensure adequate coverage of the rarest behaviors, preventing a robust per-label reliability analysis. Future data collections should increase the proportion of independently annotated material and explicitly stratify the reliability subset by behavior occurrence.
More broadly, the dataset was collected in a single daycare center with 13 participants over four days, and only 10 hourly recordings were selected for annotation. The findings should therefore not be interpreted as evidence of generalization to other care centers, cultural contexts, or IDD profiles. Broader multi-site and longitudinal data collections will be necessary to assess such generalizability.
In addition, the reported gender distribution was inferred observationally by the researchers rather than self-reported by participants. It should therefore be interpreted only as descriptive metadata and not as validated demographic information.

This work investigated body-behavior recognition in a naturalistic daycare setting involving adults with IDD. SocialMirror combines educator-informed annotation, derived representations from two camera views and a multi-view 3D branch, and individual- and relational-level recognition benchmarks. Beyond providing a research resource, the study identifies four methodological lessons for disability-inclusive affective computing: grounding recognition targets in situated knowledge, documenting behavioral diversity across individuals as well as classes, designing sensing around the joint observability of interactions, and evaluating progress through recognition performance across the behavioral repertoire.

\section*{Acknowledgments}
We would like to thank all participants and educators that took part in this study from the Fraternità\&Amicizia center.


\balance

\small
\bibliographystyle{IEEEtran}
\bibliography{__ref}

@article{DBLP:journals/mta/GelsominiSG21,
  author       = {Mirko Gelsomini and
                  Micol Spitale and
                  Franca Garzotto},
  title        = {Phygital interfaces for people with intellectual disability: an exploratory
                  study at a social care center},
  journal      = {Multim. Tools Appl.},
  volume       = {80},
  number       = {26},
  pages        = {34843--34874},
  year         = {2021},
  url          = {https://doi.org/10.1007/s11042-021-11164-9},
  doi          = {10.1007/S11042-021-11164-9},
  bibsource    = {dblp computer science bibliography, https://dblp.org}
}

@inproceedings{wittenburg2006elan,
  author    = {Peter Wittenburg and Hennie Brugman and Albert Russel and Alex Klassmann and Han Sloetjes},
  title     = {{ELAN}: a Professional Framework for Multimodality Research},
  booktitle = {Proceedings of the Fifth International Conference on Language Resources and Evaluation (LREC 2006)},
  pages     = {1556--1559},
  year      = {2006},
  address   = {Genoa, Italy},
  publisher = {European Language Resources Association (ELRA)}
}

@article{DBLP:journals/ijmms/GennariMMMMR24,
  author       = {Rosella Gennari and
                  Maristella Matera and
                  Alessandra Melonio and
                  Marco Mores and
                  Diego Morra and
                  Mehdi Rizvi},
  title        = {A rapid-prototyping toolkit for people with intellectual disabilities},
  journal      = {Int. J. Hum. Comput. Stud.},
  volume       = {192},
  pages        = {103347},
  year         = {2024},
  url          = {https://doi.org/10.1016/j.ijhcs.2024.103347},
  doi          = {10.1016/J.IJHCS.2024.103347},
  bibsource    = {dblp computer science bibliography, https://dblp.org}
}

@inproceedings{moon2023anonymization,
  title={Anonymization for skeleton action recognition},
  author={Moon, Saemi and Kim, Myeonghyeon and Qin, Zhenyue and Liu, Yang and Kim, Dongwoo},
  booktitle={Proceedings of the AAAI Conference on Artificial Intelligence},
  volume={37},
  number={12},
  pages={15028--15036},
  year={2023}
}

@article{gao2024privacy,
  title={Privacy-preserving 3-D skeleton-based video action recognition via graph convolution network},
  author={Gao, Xuesong and Li, Keqiu and Liu, Xiulong and Nie, Jie and Chen, Weiqiang and Tian, Yonghong},
  journal={IEEE Transactions on Consumer Electronics},
  volume={71},
  number={2},
  pages={6627--6641},
  year={2024},
  publisher={IEEE}
}

@inproceedings{barbareschi2026inclusive,
  title={Inclusive AI: Rethinking AI-based Multimodal Interaction for Diverse and Underrepresented Users},
  author={Barbareschi, Giulia and Kosmyna, Nataliya and Matera, Maristella and Sinha, Anoop K and Sheth, Shruti and Spitale, Micol and Vinciarelli, Alessandro},
  booktitle={Proceedings of the 28th International Conference On Multimodal Interaction},
  pages={1483--1485},
  year={2026}
}

@article{trewin2018ai,
  title={AI fairness for people with disabilities: Point of view},
  author={Trewin, Shari},
  journal={arXiv preprint arXiv:1811.10670},
  year={2018}
}

@article{moreno2024sneaky,
  title={Sneaky emotions: impact of data partitions in affective computing experiments with brain-computer interfacing},
  author={Moreno-Alcayde, Yoelvis and Traver, V Javier and Leiva, Luis A},
  journal={Biomedical Engineering Letters},
  volume={14},
  number={1},
  pages={103--113},
  year={2024},
  publisher={Springer}
}

@article{huang2026inclusive,
  title={Inclusive AI for Group Interactions: Predicting Gaze-Direction Behaviors in People with Intellectual and Developmental Disabilities},
  author={Huang, Giulia and Matera, Maristella and Spitale, Micol},
  journal={arXiv preprint arXiv:2603.14460},
  year={2026}
}

@article{hussain2024ai,
  title={AI-driven behavior biometrics framework for robust human activity recognition in surveillance systems},
  author={Hussain, Altaf and Khan, Samee Ullah and Khan, Noman and Shabaz, Mohammad and Baik, Sung Wook},
  journal={Engineering Applications of Artificial Intelligence},
  volume={127},
  pages={107218},
  year={2024},
  publisher={Elsevier}
}

@article{almufareh2024conceptual,
  title={A conceptual model for inclusive technology: advancing disability inclusion through artificial intelligence},
  author={Almufareh, Maram Fahaad and Kausar, Sumaira and Humayun, Mamoona and Tehsin, Samabia},
  journal={Journal of Disability Research},
  volume={3},
  number={1},
  pages={20230060},
  year={2024},
  publisher={King Salman Center for Disability Research}
}

@book{shneiderman2022human,
  title={Human-centered AI},
  author={Shneiderman, Ben},
  year={2022},
  publisher={Oxford University Press}
}

@article{clarke2017thematic,
  title={Thematic analysis},
  author={Clarke, Victoria and Braun, Virginia},
  journal={The journal of positive psychology},
  volume={12},
  number={3},
  pages={297--298},
  year={2017},
  publisher={Taylor \& Francis}
}

@inproceedings{tang2026disability,
  title={Disability-First AI Dataset Annotation: Co-designing Stuttered Speech Annotation Guidelines with People Who Stutter},
  author={Tang, Xinru and Li, Jingjin and Wu, Shaomei},
  booktitle={Proceedings of the 2026 CHI Conference on Human Factors in Computing Systems},
  pages={1--22},
  year={2026}
}

@article{shams2025ai,
  title={AI for All: Identifying AI incidents Related to Diversity and Inclusion},
  author={Shams, Rifat Ara and Zowghi, Didar and Bano, Muneera},
  journal={Journal of Artificial Intelligence Research},
  volume={83},
  year={2025}
}

@article{bircanin2025co,
  title={Co-designing in the middle: supporting adults with intellectual disabilities to meaningfully participate in design},
  author={Bircanin, Filip and Brereton, Margot and Sitbon, Laurianne and Ploderer, Bernd and Esteban, Michael},
  journal={CoDesign},
  volume={21},
  number={3},
  pages={607--627},
  year={2025},
  publisher={Taylor \& Francis}
}

@inproceedings{do2024skateformer,
  title={Skateformer: skeletal-temporal transformer for human action recognition},
  author={Do, Jeonghyeok and Kim, Munchurl},
  booktitle={European Conference on Computer Vision},
  pages={401--420},
  year={2024},
  organization={Springer}
}

@inproceedings{duan2022revisiting,
  title={Revisiting skeleton-based action recognition},
  author={Duan, Haodong and Zhao, Yue and Chen, Kai and Lin, Dahua and Dai, Bo},
  booktitle={Proceedings of the IEEE/CVF conference on computer vision and pattern recognition},
  pages={2969--2978},
  year={2022}
}

@inproceedings{duan2023skeletr,
  title={Skeletr: Towards skeleton-based action recognition in the wild},
  author={Duan, Haodong and Xu, Mingze and Shuai, Bing and Modolo, Davide and Tu, Zhuowen and Tighe, Joseph and Bergamo, Alessandro},
  booktitle={Proceedings of the IEEE/CVF international conference on computer vision},
  pages={13634--13644},
  year={2023}
}

@article{gupta2022human,
  title={Human activity recognition in artificial intelligence framework: a narrative review},
  author={Gupta, Neha and Gupta, Suneet K and Pathak, Rajesh K and Jain, Vanita and Rashidi, Parisa and Suri, Jasjit S},
  journal={Artificial intelligence review},
  volume={55},
  number={6},
  pages={4755--4808},
  year={2022},
  publisher={Springer}
}

@article{cibrian2025limitations,
  title={Limitations in speech recognition for young adults with down syndrome},
  author={Cibrian, Franceli L and Chen, Yingchen’Yuki’ and Anderson, Kayla and Abrahamsson, Cecilia Marie and Motti, Vivian Genaro},
  journal={Universal Access in the information Society},
  volume={24},
  number={3},
  pages={2295--2313},
  year={2025},
  publisher={Springer}
}

@article{liu2025systematic,
  title={A Systematic Review of Skeleton-Based Action Recognition: Methods, Challenges, and Future Directions},
  author={Liu, Yi and Liu, Ruyi and Hu, Yuzhi and Wu, Mengyao and Xin, Wentian and Miao, Qiguang and Wu, Shuai and Li, Long},
  journal={IEEE Transactions on Neural Networks and Learning Systems},
  year={2025},
  publisher={IEEE}
}

@article{lin2025actions,
  title={From Actions to Kinesics: Extracting Human Psychological States Through Bodily Movements},
  author={Lin, Cheyu and Flanigan, Katherine A},
  journal={arXiv preprint arXiv:2510.04844},
  year={2025}
}

@article{lasri2025dpca,
  title={DPCA-GCN: Dual-Path Cross-Attention Graph Convolutional Networks for Skeleton-Based Action Recognition},
  author={Lasri, Khadija and El Fazazy, Khalid and Mohamed Mahraz, Adnane and Tairi, Hamid and Riffi, Jamal},
  journal={Computation},
  volume={13},
  number={12},
  pages={293},
  year={2025},
  publisher={MDPI}
}

@article{panigutti2023co,
  title={Co-design of human-centered, explainable AI for clinical decision support},
  author={Panigutti, Cecilia and Beretta, Andrea and Fadda, Daniele and Giannotti, Fosca and Pedreschi, Dino and Perotti, Alan and Rinzivillo, Salvatore},
  journal={ACM Transactions on Interactive Intelligent Systems},
  volume={13},
  number={4},
  pages={1--35},
  year={2023},
  publisher={ACM New York, NY}
}

@inproceedings{shah2026axon,
  title={AXON: Action Characterization Through Cross-Modal Knowledge Distillation for Neurodiverse Individuals},
  author={Shah, Siddhant Bikram and Johnson, Kristina T},
  booktitle={Proceedings of the AAAI Conference on Artificial Intelligence},
  volume={40},
  number={46},
  pages={39228--39238},
  year={2026}
}

@inproceedings{shao2026finetec,
  title={FineTec: Fine-Grained Action Recognition Under Temporal Corruption via Skeleton Decomposition and Sequence Completion},
  author={Shao, Dian and Shi, Mingfei and Liu, Like},
  booktitle={Proceedings of the AAAI Conference on Artificial Intelligence},
  volume={40},
  number={11},
  pages={8842--8850},
  year={2026}
}

@article{silvera2025ai,
  title={AI-enabled AT Framework: a principles-based approach to emerging assistive technology},
  author={Silvera-Tawil, David and Higgins, Liesel and Packer, Katie and Bayor, Andrew A and Walker, Janine G and Li, Jane and Niven, Philippa and Khanna, Sankalp and Byrnes, Josh and Bradford, DanaKai and others},
  journal={Disability and Rehabilitation: Assistive Technology},
  volume={20},
  number={7},
  pages={1955--1974},
  year={2025},
  publisher={Taylor \& Francis}
}

@inproceedings{wu2025frequency,
  title={Frequency-semantic enhanced variational autoencoder for zero-shot skeleton-based action recognition},
  author={Wu, Wenhan and Guo, Zhishuai and Chen, Chen and Xue, Hongfei and Lu, Aidong},
  booktitle={Proceedings of the IEEE/CVF International Conference on Computer Vision},
  pages={11122--11131},
  year={2025}
}

@inproceedings{xiao2024systematic,
  title={A systematic review of ability-diverse collaboration through ability-based lens in HCI},
  author={Xiao, Lan and Bandukda, Maryam and Angerbauer, Katrin and Lin, Weiyue and Bhatnagar, Tigmanshu and Sedlmair, Michael and Holloway, Catherine},
  booktitle={Proceedings of the 2024 CHI Conference on Human Factors in Computing Systems},
  pages={1--21},
  year={2024}
}

@inproceedings{yan2018spatial,
  title={Spatial temporal graph convolutional networks for skeleton-based action recognition},
  author={Yan, Sijie and Xiong, Yuanjun and Lin, Dahua},
  booktitle={Proceedings of the AAAI conference on artificial intelligence},
  volume={32},
  number={1},
  year={2018}
}

@inproceedings{xu2024inter,
  title={Inter-x: Towards versatile human-human interaction analysis},
  author={Xu, Liang and Lv, Xintao and Yan, Yichao and Jin, Xin and Wu, Shuwen and Xu, Congsheng and Liu, Yifan and Zhou, Yizhou and Rao, Fengyun and Sheng, Xingdong and others},
  booktitle={Proceedings of the IEEE/CVF conference on computer vision and pattern recognition},
  pages={22260--22271},
  year={2024}
}

@incollection{zdravkova2022potential,
  title={The potential of artificial intelligence for assistive technology in education},
  author={Zdravkova, Katerina},
  booktitle={Handbook on intelligent techniques in the educational process: Vol 1 recent advances and case studies},
  pages={61--85},
  year={2022},
  publisher={Springer}
}

@inproceedings{duan2022poseconv3d,
  title={Revisiting Skeleton-Based Action Recognition},
  author={Duan, Haodong and Zhao, Yue and Chen, Kai and Lin, Dahua and Dai, Bo},
  booktitle={Proceedings of the IEEE/CVF Conference on Computer Vision and Pattern Recognition},
  pages={2969--2978},
  year={2022}
}

@inproceedings{multimediate2023,
  title={{MultiMediate}'23: Engagement Estimation and Bodily Behaviour Recognition in Social Interactions},
  author={M{\"u}ller, Philipp and Balazia, Michal and Baur, Tobias and Dietz, Michael and Heimerl, Alexander and Schiller, Dominik and Guermal, Mohammed and Thomas, Dominike and Br{\'e}mond, Fran{\c{c}}ois and Alexandersson, Jan and Andr{\'e}, Elisabeth and Bulling, Andreas},
  booktitle={Proceedings of the 31st ACM International Conference on Multimedia},
  pages={9640--9645},
  year={2023},
  doi={10.1145/3581783.3613851}
}

@inproceedings{conflab2022,
  title={ConfLab: A Data Collection Concept, Dataset, and Benchmark for Machine Analysis of Free-Standing Social Interactions in the Wild},
  author={Raman, Chirag and Vargas Quiros, Jose and Tan, Stephanie and Islam, Ashraful and Gedik, Ekin and Hung, Hayley},
  booktitle={Advances in Neural Information Processing Systems},
  volume={35},
  year={2022}
}

@inproceedings{chen20243dsa,
  title={3DSA: Multi-View 3D Human Pose Estimation With 3D Space Attention Mechanisms},
  author={Chen, Pohan and Tsi, Chiachi},
  booktitle={European Conference on Computer Vision},
  year={2024}
}

@inproceedings{liao2024mvgformer,
  title={Multiple View Geometry Transformers for 3D Human Pose Estimation},
  author={Liao, Ziwei and Zhu, Jialiang and Wang, Chunyu and Hu, Han and Waslander, Steven L.},
  booktitle={Proceedings of the IEEE/CVF Conference on Computer Vision and Pattern Recognition},
  year={2024}
}

@inproceedings{li2023mmasd,
  title={Mmasd: A multimodal dataset for autism intervention analysis},
  author={Li, Jicheng and Chheang, Vuthea and Kullu, Pinar and Brignac, Eli and Guo, Zhang and Bhat, Anjana and Barner, Kenneth E and Barmaki, Roghayeh Leila},
  booktitle={Proceedings of the 25th International Conference on Multimodal Interaction},
  pages={397--405},
  year={2023}
}

@inproceedings{garcia2026summary,
  title={Summary of the unusual activity recognition challenge for developmental disability support},
  author={Garcia, Christina and Le, Nhat Tan and Fujioka, Taihei and Dobhal, Umang and Shoumi, Milyun Ni’ma and Nguyen, Thanh Nha and Inoue, Sozo},
  booktitle={Journal of Physics: Conference Series},
  volume={3180},
  number={1},
  pages={012020},
  year={2026},
  organization={IOP Publishing}
}

{\appendices

\section{Thematic Analysis of Educator Interviews}
\label{app:thematic_analysis}

Table~\ref{tab:thematic_analysis} summarizes the main themes and design insights that emerged from the educator interviews and informed the development of the SocialMirror annotation framework.

\begin{table*}[htbp]
    \centering
    \caption{Thematic Analysis of Educator Interviews: Needs and Design Insights}
    \label{tab:thematic_analysis}
    \small
    \begin{tabularx}{\textwidth}{
        >{\raggedright\arraybackslash}p{3.5cm}
        >{\raggedright\arraybackslash}p{3.5cm}
        X}
        \toprule
        \textbf{Themes} & \textbf{Codes} & \textbf{Description \& Quotes} \\
        \midrule

        \textbf{Emotional \& Relational Dynamics}
        & Emotion Recognition; Emotional Contagion; Relational Autonomy
        & Educators emphasized that true independence involves ``relational autonomy,'' which requires recognizing both one's own emotions and those of others to prevent the spread of distress.
        \par \textit{(P3, Interview 1): ``Autonomy is not about knowing how to run the washing machine... it is relational autonomy, and relational autonomy must recognize your own emotions and those of your peer.''}
        \par \textit{(P3, Interview 1): Mentioned the risk of an ``uncontrolled spread of emotional contagion.''} \\

        \addlinespace

        \textbf{Personalization to User Specifics}
        & Individual Variability; Subject-specific Calibration; Rejecting Broad Labels
        & The technology must adapt to the unique history and behavioral traits of each individual rather than relying on broad clinical diagnoses.
        \par \textit{(P1, Interview 1): ``We should always start from people's stories... not from the diagnosis.''}
        \par \textit{(P2, Interview 2): ``I would like to have a personalized artificial intelligence... so that if a new person enters the center... the system could be personalized.''} \\

        \addlinespace

        \textbf{Non-Diagnostic ``Reading Reality''}
        & Human-in-the-Loop; Objective Observation; Real-time Alerting
        & The AI system should act as a supportive tool that provides objective observations without attempting to replace human clinical judgment.
        \par \textit{(P2, Interview 2): Validated the tool as ``tools for reading reality... for reading a fact.''} \\

        \addlinespace

        \textbf{Unsupervised Context Monitoring}
        & Unstructured Socialization; Post-lunch Breaks; Privacy Constraints
        & Monitoring is considered most valuable during unstructured social moments where direct supervision is lower, such as post-lunch breaks, while balancing privacy.
        \par \textit{(P2, Interview 2): ``During the post-lunch break, for example... the operator would step away... and the attendees are on their own... there is always an operator moving through the corridors, and they could be helped by this type of device.''}
        \par \textit{(P1, Interview 2): ``We struggle a bit... due to the privacy issue, because then everyone enters.''} \\

        \bottomrule
    \end{tabularx}
\end{table*}


\section{Additional Benchmark Diagnostics}
\label{app:benchmark_diagnostics}

The class-level results expose the strong effect of label imbalance.
ClosedPosture is recognized substantially more reliably than the
minority classes in both monocular views, with F1 scores of 0.830
and 0.850 for \textit{cam1} and \textit{cam2}, respectively. In contrast,
DisengagedRepetitive remains particularly challenging, with F1
scores of 0.026 and 0.037. The 3D baseline exhibits a different
error profile: Stand and ClosedPosture are partially discriminated,
whereas Sit is not correctly recognized in the evaluated holdout
subset. These results further motivate the use of macro-averaged
metrics rather than raw accuracy for benchmark interpretation.

\begin{table*}[htbp]
    \centering
    \caption{Per-class precision, recall, and F1 scores for the reported SVM baselines. \textit{Cam1} and \textit{cam2} results correspond to the video-level grouped evaluation, whereas the 3D results correspond to the fixed holdout evaluation.}
    \label{tab:per_class_results}
    \small
    \begin{tabular}{llccc}
        \toprule
        \textbf{Setting} & \textbf{Class} & \textbf{Precision} & \textbf{Recall} & \textbf{F1} \\
        \midrule

        \textit{cam1}
        & Stand & 0.367 & 0.225 & 0.279 \\
        & Sit & 0.116 & 0.129 & 0.122 \\
        & ClosedPosture & 0.834 & 0.826 & 0.830 \\
        & DisengagedRepetitive & 0.024 & 0.028 & 0.026 \\

        \midrule

        \textit{cam2}
        & Stand & 0.068 & 0.355 & 0.114 \\
        & Sit & 0.393 & 0.218 & 0.281 \\
        & ClosedPosture & 0.944 & 0.774 & 0.850 \\
        & DisengagedRepetitive & 0.026 & 0.060 & 0.037 \\

        \midrule

        3D
        & Stand & 0.556 & 0.455 & 0.500 \\
        & Sit & 0.000 & 0.000 & 0.000 \\
        & ClosedPosture & 0.519 & 0.932 & 0.667 \\
        & DisengagedRepetitive & 0.400 & 0.133 & 0.200 \\

        \bottomrule
    \end{tabular}
\end{table*}

\begin{table*}[htbp]
    \centering
    \caption{Confusion matrices for the \textit{cam1}, \textit{cam2}, and 3D SVM baselines. Rows correspond to ground-truth classes and columns to predicted classes. St = Stand, Si = Sit, CP = ClosedPosture, and DR = DisengagedRepetitive.}
    \label{tab:confusion_matrices}
    \small

    \begin{minipage}[t]{0.32\textwidth}
        \centering
        \textbf{(a) \textit{cam1}}\\[3pt]
        \begin{tabular}{lrrrr}
            \toprule
             & St & Si & CP & DR \\
            \midrule
            St & 131 & 48 & 283 & 120 \\
            Si & 108 & 83 & 322 & 132 \\
            CP & 38 & 519 & 11741 & 1918 \\
            DR & 80 & 68 & 1727 & 54 \\
            \bottomrule
        \end{tabular}
    \end{minipage}
    \hfill
    \begin{minipage}[t]{0.32\textwidth}
        \centering
        \textbf{(b) \textit{cam2}}\\[3pt]
        \begin{tabular}{lrrrr}
            \toprule
             & St & Si & CP & DR \\
            \midrule
            St & 166 & 27 & 239 & 36 \\
            Si & 108 & 112 & 234 & 59 \\
            CP & 2131 & 114 & 8126 & 128 \\
            DR & 50 & 32 & 12 & 6 \\
            \bottomrule
        \end{tabular}
    \end{minipage}
    \hfill
    \begin{minipage}[t]{0.32\textwidth}
        \centering
        \textbf{(c) 3D}\\[3pt]
        \begin{tabular}{lrrrr}
            \toprule
             & St & Si & CP & DR \\
            \midrule
            St & 20 & 0 & 22 & 2 \\
            Si & 6 & 0 & 12 & 1 \\
            CP & 1 & 0 & 55 & 3 \\
            DR & 9 & 0 & 17 & 4 \\
            \bottomrule
        \end{tabular}
    \end{minipage}

\end{table*}

\subsection{Participant-Level Label Distribution}

Table~\ref{tab:participant_distribution} summarizes the participant
coverage of each target class. The distributions reveal substantial
participant--label concentration, particularly for minority behaviors.
For example, DisengagedRepetitive is represented by only three
participants in \textit{cam1}, one participant in \textit{cam2}, and two
participants in the 3D subset. This further motivates interpreting the
reported evaluation as cross-video rather than participant-independent
generalization.

\begin{table*}[htbp]
    \centering
    \caption{Participant-level distribution of the target classes across the three benchmark settings. \#Part. denotes the number of participants contributing at least one window to the corresponding class. Top share denotes the proportion of class windows contributed by the participant with the largest number of samples.}
    \label{tab:participant_distribution}
    \small
    \begin{tabular}{llrrrr}
        \toprule
        \textbf{Setting} & \textbf{Class} & \textbf{Windows} & \textbf{\#Part.} & \textbf{Top participant} & \textbf{Top share} \\
        \midrule

        \textit{cam1}
        & Stand & 582 & 11 & P01 & 37.63\% \\
        & Sit & 645 & 11 & P01 & 35.66\% \\
        & ClosedPosture & 14216 & 8 & P02 & 49.98\% \\
        & DisengagedRepetitive & 1929 & 3 & P07 & 62.67\% \\

        \midrule

        \textit{cam2}
        & Stand & 468 & 8 & P01 & 45.09\% \\
        & Sit & 513 & 8 & P01 & 44.05\% \\
        & ClosedPosture & 10499 & 7 & P02 & 68.56\% \\
        & DisengagedRepetitive & 100 & 1 & P01 & 100.00\% \\

        \midrule

        3D
        & Stand & 176 & 9 & P01 & 65.34\% \\
        & Sit & 75 & 8 & P01 & 36.00\% \\
        & ClosedPosture & 236 & 6 & P05 & 38.14\% \\
        & DisengagedRepetitive & 120 & 2 & P01 & 75.83\% \\

        \bottomrule
    \end{tabular}
\end{table*}

\clearpage
\section{Full Annotation Codebook}
\label{app:full_codebook}
\begin{table*}[htb!]
    \centering
    \caption{Summary of the complete annotation codebook. All categories describe directly observable behaviors, interactions, contextual interventions, or annotation-quality conditions rather than clinical or diagnostic states.}
    \label{tab:full_codebook}
    \small

    \begin{tabularx}{\textwidth}{
        >{\raggedright\arraybackslash}p{2.4cm}
        >{\centering\arraybackslash}p{0.9cm}
        >{\raggedright\arraybackslash}p{3.2cm}
        >{\raggedright\arraybackslash}X}
        \toprule
        \textbf{Area} & \textbf{Code} & \textbf{Label} & \textbf{Operational description} \\
        \midrule

        \textbf{Individual}
        & A1a
        & Stand
        & Transition from a sitting to a standing position. \\

        & A1b
        & Sit
        & Transition from a standing to a sitting position. \\

        & A2a
        & Enter
        & Person enters the main observed activity area. \\

        & A2b
        & Exit
        & Person leaves the main observed activity area. \\

        & A3
        & ClosedPosture
        & Inward-oriented or reduced-open body posture maintained for at least 5\,s. \\

        & A4
        & DisengagedBehaviour
        & Observable disengagement or stable orientation away from the group or ongoing activity for at least 5\,s. \\

        & A5
        & RepetitiveMovement
        & Repeated self-similar movement, defined as at least three repetitions within 10\,s or persistence for at least 5\,s. \\

        \midrule

        \textbf{Relational}
        & B1
        & Proximity
        & Sustained, non-accidental physical proximity between two people for at least 5\,s. \\

        & B2
        & AffiliativeContact
        & Observable positive or affiliative physical contact between two people. \\

        & B3
        & AmbiguousContact
        & Physical contact whose relational meaning cannot be determined reliably from observation. \\

        & B4
        & MutualOrientation
        & Two people observably oriented toward one another for at least 2\,s. \\

        & B5
        & PossibleTension
        & Observable interactional configuration potentially consistent with tension, without clinical interpretation. \\

        & B6
        & AttentionReaction
        & Observable reaction by one person to another person's attentional cue or action. \\

        \midrule

        \textbf{Support context}
        & C1
        & Support
        & Observable educator support directed toward a person or ongoing situation. \\

        & C2
        & Co-regulation
        & Observable educator intervention aimed at jointly regulating an ongoing situation or behavior. \\

        \midrule

        \textbf{Tag notes}
        & D1
        & PoorVisibility
        & Visibility is insufficient for reliable behavioral interpretation, for example because of occlusion. \\

        & D2
        & AmbiguousMeaning
        & The behavior is visible, but its meaning cannot be determined reliably. \\

        & D3
        & PossibleDiscomfort
        & Observable cue potentially consistent with discomfort, recorded descriptively and without clinical interpretation. \\

        \bottomrule
    \end{tabularx}
\end{table*}}

\end{document}